\documentclass[10pt,twocolumn,letterpaper]{article}
\usepackage{mytitlesec}
\usepackage[pagenumbers]{cvpr}   

\definecolor{cvprblue}{rgb}{0.21,0.49,0.74}
\usepackage[pagebackref,breaklinks,colorlinks,allcolors=cvprblue]{hyperref}

\usepackage{stfloats}
\usepackage{hyperref}
\usepackage{url}
\usepackage{graphicx}
\usepackage{wrapfig}
\usepackage{float}
\usepackage{booktabs}
\usepackage{multirow}
\usepackage{xcolor}    
\usepackage{makecell}
\usepackage{cuted}     
\usepackage{caption}   
\usepackage{subcaption}   
\usepackage{adjustbox}    
\usepackage[table]{xcolor}
\definecolor{slateL}{HTML}{F3F5F7} 
\definecolor{slateM}{HTML}{E5EAF0} 
\definecolor{slateH}{HTML}{D0D7E3} 

\colorlet{lightyellow}{slateL}
\colorlet{orange}{slateM}
\colorlet{tablered}{slateH}

\titlespacing\section{0pt}{0pt plus 0pt minus 0pt}{0pt plus 0pt minus 0pt}
\titlespacing\subsection{0pt}{0pt plus 0pt minus 0pt}{0pt plus 0pt minus 0pt}
\titlespacing\subsubsection{0pt}{0pt plus 0pt minus 0pt}{0pt plus 0pt minus 0pt}

\usepackage{tikz}

\title{Spatial Retrieval Augmented Autonomous Driving}

\author{Xiaosong Jia$^{1\ast}$, Chenhe Zhang$^{1\ast}$, Yule Jiang$^{2\ast}$, Songbur Wong$^{2\ast}$, \\
Zhiyuan Zhang$^2$, Chen Chen$^{3}$, Shaofeng Zhang$^{4}$, Xuanhe Zhou$^2$, Xue Yang$^{2\dagger}$, \\ Junchi Yan$^{2\dagger}$, Yu-Gang Jiang$^{1}$ \\
[1mm]
1. Institute of Trustworthy Embodied AI, Fudan University \\
2. Shanghai Jiao Tong University  \\
3. Key Laboratory of Target Cognition and Application Technology, \\ 
Aerospace Information Research Institute, Chinese Academy of Sciences \\
4. University of Science and Technology of China \\
\normalsize{
$^\ast$Equal contributions \quad $^\dagger$Correspondence authors}
\\
\normalsize{
\url{https://spatialretrievalad.github.io/}
}
\vspace{-5mm}
}

\usepackage[dvipsnames]{xcolor}
\definecolor{red}{HTML}{2A69AC}  
\definecolor{blue}{HTML}{329BA7}
\definecolor{teak}{RGB}{181,114,33}

\usepackage{docmute}

\begin{document}
\maketitle

\begin{strip}
\centering
\includegraphics[width=0.92\textwidth]{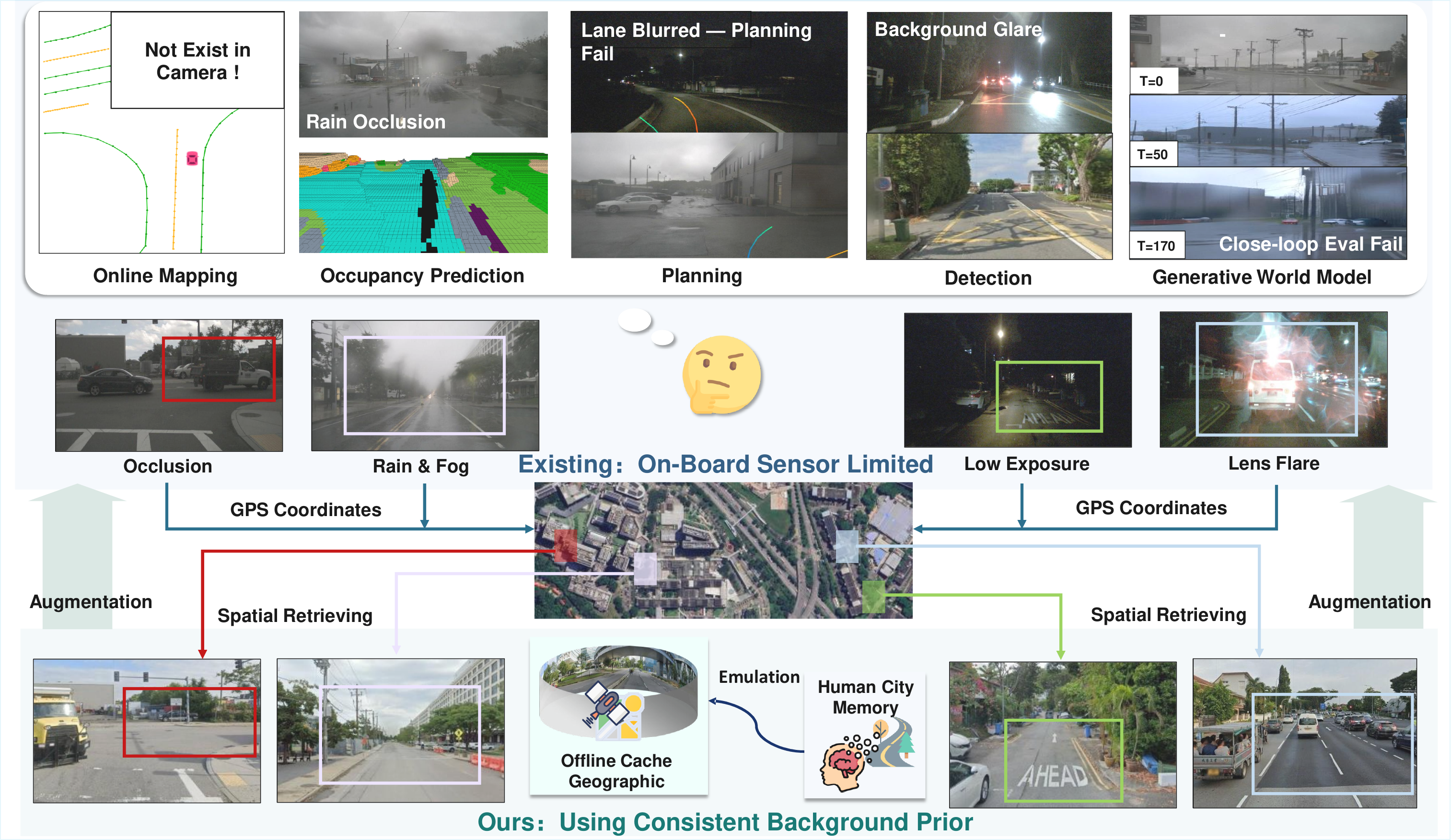}
\captionof{figure}{Existing autonomous driving systems (upper) largely rely on onboard sensors, which are vulnerable to perceptual conditions. Inspired by human drivers’ ability to recall memory of previously seen roads, we introduce the spatial retrieval paradigm (lower) which utilizes offline cached geographic images as an extra input modality to enhance the performance of multiple AD tasks.}
\label{fig:teaser}
\end{strip}

\begin{abstract}

Existing autonomous driving systems rely on onboard sensors (cameras, LiDAR, IMU, etc) for environmental perception. However, this paradigm is limited by the drive-time perception horizon and often fails under limited view scope, occlusion or extreme conditions such as darkness and rain. In contrast, human drivers are able to recall road structure even under poor visibility. To endow models with this ``recall" ability, we propose the spatial retrieval paradigm, introducing offline retrieved geographic images as an additional input. These images are easy to obtain from offline caches (e.g, Google Maps or stored autonomous driving datasets) without requiring additional sensors, making it a plug-and-play extension for existing AD tasks.

For experiments, we first extend the nuScenes dataset with geographic images retrieved via Google Maps APIs and align the new data with ego-vehicle trajectories. We establish baselines across five core autonomous driving tasks: object detection, online mapping, occupancy prediction, end-to-end planning, and generative world modeling.  Extensive experiments show that the extended modality could enhance the performance of certain tasks. We will open-source dataset curation code, data, and benchmarks for further study of this new autonomous driving paradigm.

\end{abstract}
\section{Introduction}


Modern autonomous driving (AD) methods rely on onboard sensors, including cameras, LiDAR, and IMU, to capture environmental information~\cite{liu2022bevfusion,chitta2022transfuser}. While this paradigm already achieves strong performance~\cite{hu2023_uniad,wu2022trajectory,li2025think2drive,hu2023planning}, their inputs are \textbf{inherently constrained by the limited range and line-of-sight nature of online sensing}. As a result, in visually challenging scenarios such as limited view scope, occlusion, extreme exposure, or adverse weather conditions (rain, snow, fog, etc),  the system performance can consequently degrade significantly ~\cite{zhang2023perception,li2023delving}, as in Fig.~\ref{fig:teaser}. For instance, tasks like online mapping~\cite{liao2022maptr,onlinemapvectorization,Yuan_2024_streammapnet,liu2023vectormapnet} and occupancy prediction~\cite{yu2023flashocc,li2023fb,chen2025sa,tian2023occ3d,huang2023tri} aim to estimate scene structure and limited visibility or occlusions can degrade their recognition of environment and subsequently impair planning~\cite{jiang2024robuste2e}. Similarly, recent AD world models~\cite{gao2024magicdrivedit,unimlvg,bench2driveR} struggle to generate novel scenes when the ego-vehicle deviates significantly from recorded logs~\cite{you2024bench2drive,wang2024freevs}, a limitation tied to the small scope of onboard views, 
restricting their capability to serve as a simulator for closed-loop evaluation and reinforcement learning.

On the contrary, human drivers recall recent memory of the scene when current visual input is insufficient~\cite{luck1997capacity}. In this work, \textbf{we aim to enhance AD stacks with broader context beyond the immediate range of onboard sensors by spatial retrieval}. The spatial geographic data could be from Google Maps, where street view and satellite images with latitude and longitude are provided. For autonomous driving companies, their offline cached dataset could also be used. Unlike onboard sensors~\cite{caesar2020nuscenes,kitti,Bench2Drive,carla}, these geographic data are offline, globally accessible, and unaffected by drive-time disturbances. They provide rich contextual cues from perspectives beyond the ego vehicle, offering a cost-effective way to enrich spatial context without additional sensors or manual annotations.

To systematically investigate this new paradigm, we first build a framework to integrate geographic data into existing autonomous driving datasets. The framework automates data collection and spatial alignment via Google Maps APIs~\cite{googleearth} and ego-pose information to fetch and align coordinate systems.  Using this framework, we then extend the nuScenes dataset~\cite{caesar2020nuscenes} with corresponding geographic images and coordinate-based spatial retrieval APIs. Finally, to study the effects of this new modality, we establish baselines across five key AD tasks: object detection, online mapping, occupancy prediction, end-to-end planning, and generative world modeling. We design a plug-and-play adapter to seamlessly integrate the geographic images into existing models. \textbf{Extensive experiments demonstrate this modality improves performance across  tasks}.

Our main contributions are summarized as follows:
\begin{itemize}
\item We introduce the \textbf{spatial retrieval paradigm} for AD, which mitigates the environmental sensitivity of onboard perception and supplies broad, far-horizon context.
\item We construct an \textbf{extended nuScenes dataset} - nuScenes-Geography with geographic images and spatial retrieval APIs, enabling systematic study of the  new paradigm.
\item We design a \textbf{model-agnostic adapter} and establish baselines across five  AD tasks, demonstrating the broad applicability provided by the new modality.
\item We will \textbf{open-source} data curation pipelines, extended dataset, and all baselines to facilitate future research.
\end{itemize}

\section{Related Work}

\noindent\textbf{Autonomous Driving Paradigms.}  Despite recent advances in end-to-end pipelines~\cite{hu2023_uniad,jiang2023vad,sun2025sparsedrive, weng2024drive, drivetransformer, yang2022learning, jia2023hdgt,yang2311llm4drive,jia2023driveadapter,jia2024bench2drive}, multisensor fusion~\cite{fadadu2022multi, liu2022bevfusion, liang2022bevfusion, prakash2021multi, chitta2022transfuser, cai2023bevfusion4d,jia2025drivetransformer,zhu2024flatfusion,zhang2025pointobb}, and temporal modeling~\cite{li2022bevformer,wang2023exploring,park2022time,lin2023sparse4d,yuan2024streammapnet,yang2025drivemoe}, these methods are constrained by limited drive-time perception horizon, lacking long-range, location-specific priors. For offline generative world models, existing methods~\cite{vista,unimlvg,gao2023magicdrive,li2024drivingdiffusion,lu2024wovogen,yang2025resim} are prone to hallucinate when driving deviation is large due to the lack of spatial grounding. The proposed geographic image modality offers strong spatial information.


\noindent\textbf{Retrieval Methods in Autonomous Driving.} Retrieval has been used for visual place recognition~\cite{wang2025rad,wang2024rac3}, rule understanding with LLM~\cite{cai2024driving}, localization~\cite{hu2024combining}, and planning trajectories sampling~\cite{casas2021mp3} in autonomous driving. The proposed spatial retrieval paradigm instead fetches location-aligned geographic images and treats them as complementary sensory inputs of the AD system.

\begin{table}[!t]
\centering
\small
\setlength{\tabcolsep}{4pt}
\caption{\textbf{Evaluated Tasks for Spatial Retrieval Paradigm.}}
\vspace{-3mm}
\begin{tabular}{lll}
\toprule
\textbf{Task} & \textbf{Usage} & \textbf{Output} \\
\midrule
Detection & On-Vehicle & 3D Bounding Boxes \\
Mapping & On-Vehicle & Road Lines \\
Occupancy & On-Vehicle & 3D Occupancy Grid \\
Planning & On-Vehicle & Ego Trajectory \\
World Model & Offline Server & Future Video Frames \\
\bottomrule
\end{tabular}
\label{tab:tasks}
\vspace{-4mm}
\end{table}

\section{Methodology}

\subsection{Spatial Retrieval Paradigm \& Task Definition}

Suppose a clip of autonomous driving data $\mathcal{D}_{AD}$ is composed of $T$ time-sequenced sensor and pose data $\{(S_t, P_t)\}_{t=1}^T$, where each time-step $t$ has   onboard sensor data $S_t$ (e.g., multi-view images $I_t$ with camera intrinsics and extrinsics) and the ego-vehicle's pose $P_t$. 

An offline geographic database $\mathcal{D}_{geo}$ is introduced, composed of geographic images $I_{geo}$ and their corresponding metadata (global coordinates and camera parameters) $P_{geo}$.

We evaluate the effectiveness of the spatial retrieval paradigm across five AD tasks (Table~\ref{tab:tasks}). For on-vehicle tasks (3D object detection, online mapping, occupancy prediction, and motion planning), a retrieval function $\mathcal{R}$ is applied at each time-step $t$, where the function takes current images $I_t$ and ego-pose $P_t$ as inputs and retrieves from $\mathcal{D}_{geo}$ to fetch the most relevant geographic data:
\begin{equation}
\setlength{\abovedisplayskip}{0pt}
\setlength{\belowdisplayskip}{0pt}
    \text{RetrievedGeoData}_t = \mathcal{R}(I_t, I_{geo}, P_t, P_{geo})
\end{equation}

In this work, for simplicity, we retrieve the nearest geographic images for each camera at each time-step. If the 3D distance is larger than a threshold, the API returns NONE. There could be more advanced ways of retrieval, for example, more retrieved neighborhood images as global context, which we leave for future works to explore.

For the offline task (generative world model), multiple geographic images are retrieved along the desired driving trajectories of the generation, which provides a  spatial scaffold to guide long-horizon, globally consistent scene generation with less hallucination.

\begin{figure*}[t!]
    \centering
    \includegraphics[width=\textwidth]{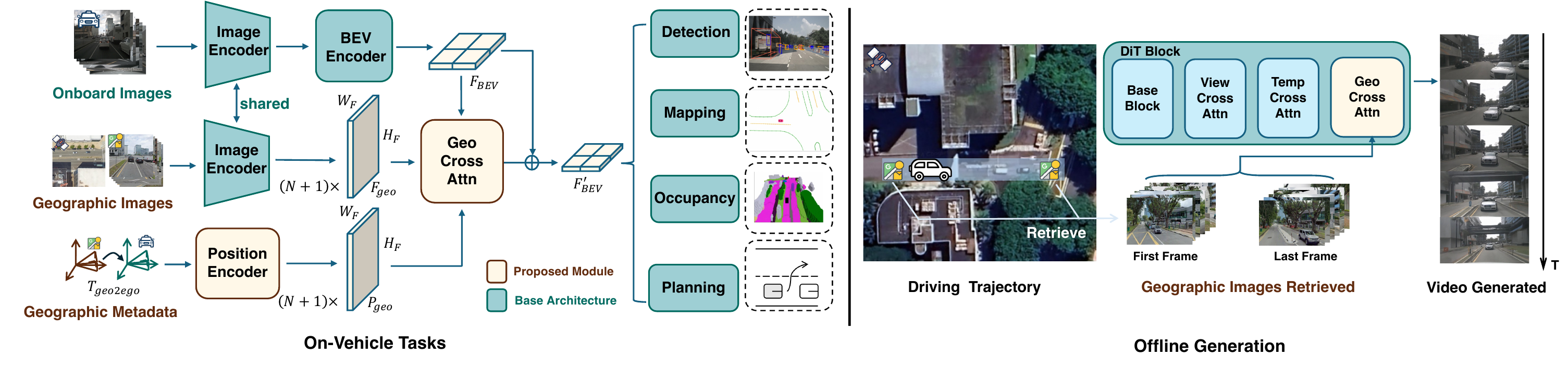} 
    \vspace{-6mm}
    \caption{\textbf{Spatial Retrieval Adapter.} We adopt cross-attention with standard BEV features ($\mathbf{F}_{\text{BEV}}$) as query and the retrieved geographic features ($\mathbf{F}_{\text{geo}}$) plus corresponding 3D positional encodings ($\mathbf{P}_{\text{geo}})$ as key and value. For generative world model task, we use similar cross-attention architecture, with the noised latents as query. Further, since the whole driving trajectory is known for video diffusion, we retrieve the corresponding geographic images based on the start and end frame positions so that the background becomes consistent.}
   \vspace{-2mm}
    \label{fig:method} 
\end{figure*}

\subsection{Spatial Retrieval Adapter}

In this section, we introduce  a general plug-and-play module, as in Fig.~\ref{fig:method} (Left), to incorporate the retrieved geographic data for BEV-based on-vehicle tasks, serving as an intuitive baseline. More advanced modules combining priors of each individual task are left for future works.

\noindent \textbf{Geographic Image and Positional Encoding.}
The retrieved geographic images are encoded by the same backbone used for onboard cameras to obtain $\mathbf{F}_{\text{geo}}$. To encode the relative spatial relationship between retrieved geographic images and current ego position,  we adopt PETR~\cite{liu2022petr} to encode their geographic image patches' 3D positional encoding to obtain $\mathbf{F}^{\text{pos}}_{\text{geo}}$.

\noindent \textbf{Geo-Cross-Attention.}
Geographic features are integrated into BEV representations via cross-attention with positional encoding, modulated by the reliability score $w$ to handle miss or wrong retrieval (detailed in Section~\ref{sec:adaptive_fusion}):
\begin{equation}
\setlength{\abovedisplayskip}{3pt}
\setlength{\belowdisplayskip}{3pt}
\mathbf{F}_{\text{BEV}}' = \mathbf{F}_{\text{BEV}} + w \cdot \text{CrossAttn}(\mathbf{F}_{\text{BEV}}, \mathbf{F}_{\text{geo}} + \mathbf{F}^{\text{pos}}_{\text{geo}}, \mathbf{F}_{\text{geo}}),
\end{equation}
The augmented BEV features are then fed to original downstream task heads. This plug-and-play design keeps all training objectives and network architectures unchanged.

\begin{figure}[t!]
    \centering
    \includegraphics[width=0.38\textwidth]{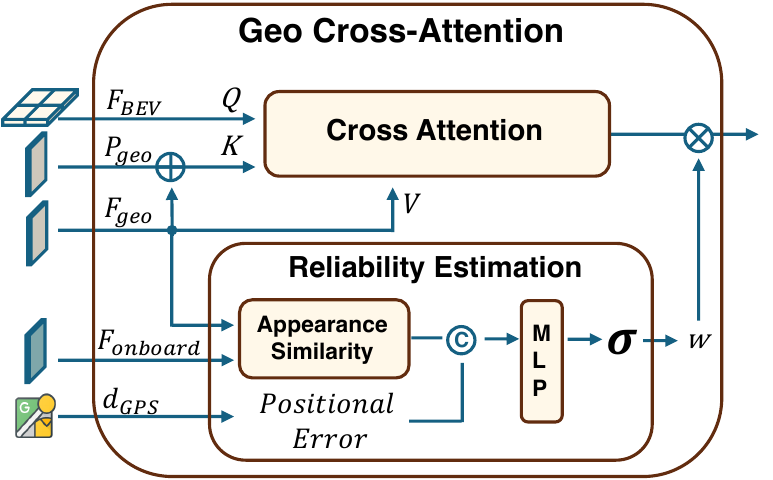} 
    \caption{\textbf{Reliability Estimation Gate.}    
    We set a Reliability Estimation Gate: when retrieved geographic missing or misaligned, the residual update approximates zero based on difference between pose and image feature.}
    \vspace{-4mm}
    \label{fig:gate} 
\end{figure}

\begin{figure*}[!t]
    \centering
    \includegraphics[width=1.0\textwidth]{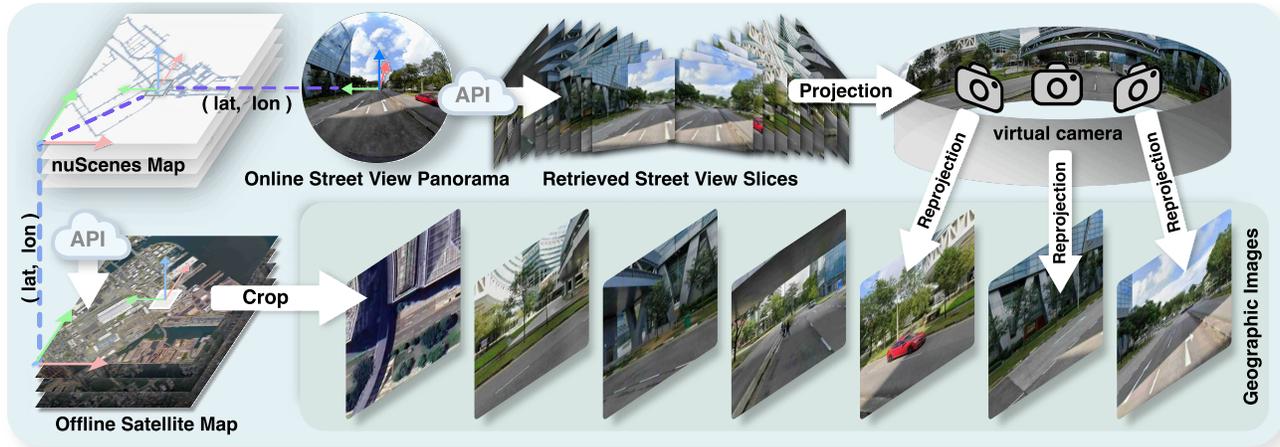} 
    \vspace{-4mm} 
    \caption{
    \textbf{Geographic Data Curation from Google Maps}. We use the GPS coordinates from nuScenes ego poses to query Google Map. Each unique panorama is downloaded once, decomposed into 18 yaw-sampled views, and projected onto an equirectangular panorama representation. For each camera at each frame in nuScenes, a virtual camera with matched intrinsics/extrinsics reprojects an aligned street view image from equirectangular panorama, effectively reducing redundant downloads and storage.
    }
    \vspace{-4mm}
    \label{fig:dataset-construction} 
\end{figure*}

\subsection{Spatial Retrieval For Generative World Model}

World models~\cite{wang2023drive,wang2024freevs,bench2driveR,yang2025resim} for autonomous driving could serve as data generator, closed-loop evaluator, or RL environment, where they usually run on clusters and servers instead of on-vehicle. As a result, these models have access to future ego trajectory, which allows to pre-fetch geographic images along upcoming path, similar to Bench2Drive-R~\cite{bench2driveR}. By injecting geographic images at future positions throughout generation, they provide persistent spatial cues that maintain scene consistency.

\noindent \textbf{Geography Extended DiT.}
Similar to Bench2Drive-R~\cite{bench2driveR}, to incorporate geographic data into generation, we inject an extra geographic cross-attention layer into the widely used DiT~\cite{peebles2023scalable} block after original attention layers:
\begin{equation}
\setlength{\abovedisplayskip}{3pt}
\setlength{\belowdisplayskip}{3pt}
\mathbf{F}^\prime = \mathbf{F} + w \cdot \text{CrossAttn}(\mathbf{F}, \mathbf{F}_{\text{geo}}+\mathbf{F}^{\text{pos}}_{\text{geo}}, \mathbf{F}_{\text{geo}}),
\end{equation}
where $\mathbf{F}$ denotes noised latent feature and \(\mathbf{F}_{\text{geo}}\) denotes  retrieved geographic features for the start and end frames of the generated segment. This design allows the model to have geographic context at corresponding future locations.

\subsection{Adaptive Fusion with Reliability Estimation}
\label{sec:adaptive_fusion}
\textbf{A key challenge in leveraging geographic data is handling missing or misaligned street view images}, as in Fig.~\ref{fig:failure_case}. To reduce the influence of these cases and let the model be robust to unreliable retrieval, we introduce an adaptive fusion mechanism, as in Fig.~\ref{fig:gate}, that dynamically adjusts the contribution of geographic features based on (i) the distance between retrieved location and ego pose; (ii) the similarity of  retrieved and current on-board images.

Specifically, we set a Reliability Estimation gate module to output reliability score $w \in [0,1]$:
\begin{equation}
\setlength{\abovedisplayskip}{3pt}
\setlength{\belowdisplayskip}{3pt}
w = \sigma(\text{MLP}([\text{ZNCC}(\mathbf{F}_{\text{onboard}}, \mathbf{F}_{\text{geo}}), d_{\text{GPS}}])),
\end{equation}
where ZNCC computes the zero-normalized cross-correlation~\cite{lewis1995fast} between onboard and geographic features, $d_{\text{GPS}}$ is the distance between the street view location and ego position, and $\sigma$ is the sigmoid function. During training, we supervise $w$ with binary labels (0 for invalid/missing, 1 for valid). At test time, this learned estimator can down-weight unreliable geographic features.

\begin{figure}[t!]
    \centering
    \includegraphics[width=0.5\textwidth]{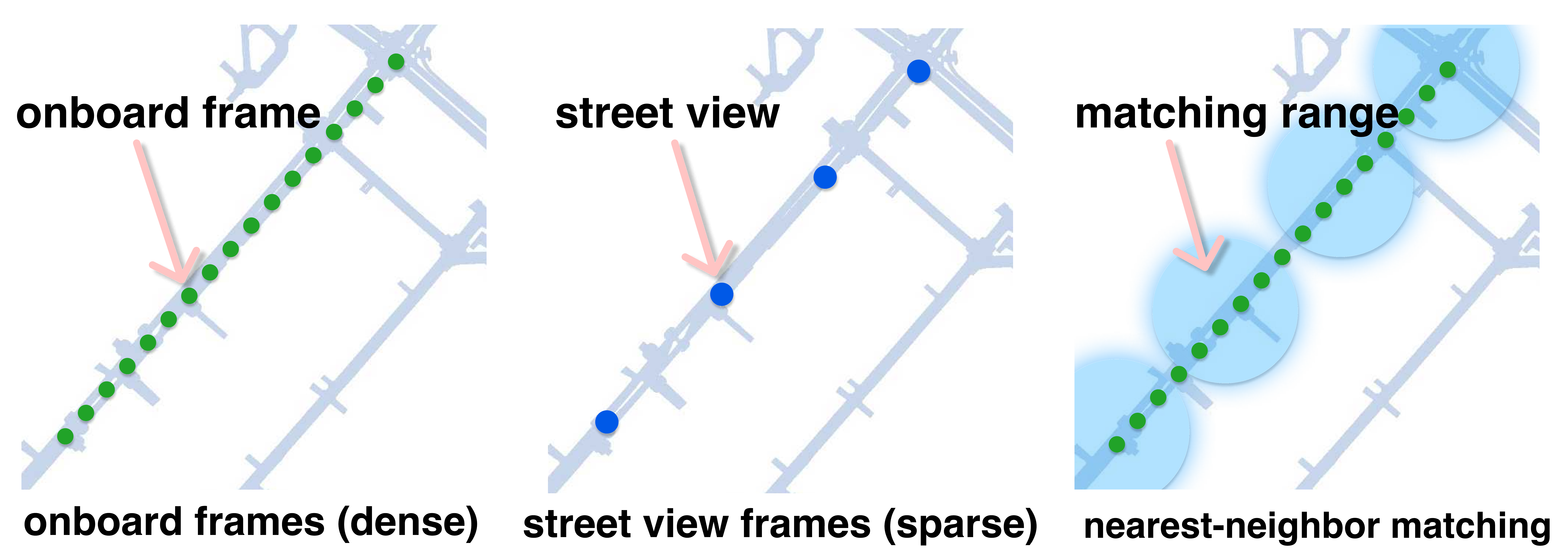}
    \vspace{-5mm}
    \caption{
        \textbf{Correspondence Relation between Frames and Geographic Data.} nuScenes frames exhibit a higher acquisition frequency than street view data along the same road. Each frame is matched to its geographically nearest street view data, where multiple frames may correspond to the same street view data.
    }
    \vspace{-5mm}
    \label{fig:dataset-matching} 
\end{figure}

\label{sec:data}

\section{nuScenes-Geography: Extended Geographic Data from Google Maps}

To systematically investigate the effectiveness of the spatial retrieval paradigm, we introduce \textbf{nuScenes-Geography}, extending the widely used nuScenes~\cite{caesar2020nuscenes} dataset with geographic data. We collect data using the Google Maps APIs, which provide access to perspective views and satellite images given coordinates and camera parameters, as in Fig.~\ref{fig:dataset-construction}.  


\noindent \textbf{Coordinate Computation.}
To associate geographic data with nuScenes frames, we compute the longitude and latitude coordinates of each frame by combining the global origin of the nuScenes map with the ego-vehicle pose. Using these coordinates, we query the Google Maps APIs to obtain street view images and satellite map tiles.

\begin{figure*}[t!]
    \centering
    \includegraphics[width=1.0\textwidth]{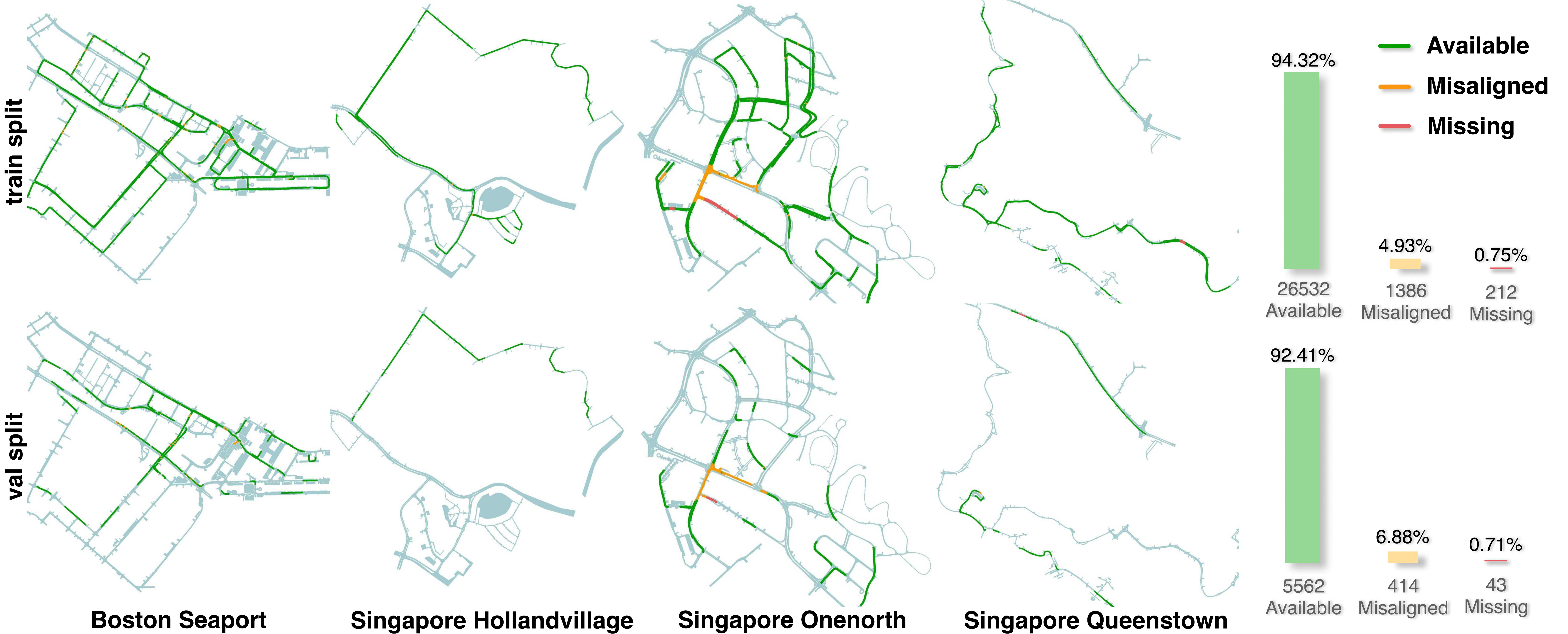} 
    \caption{
\textbf{Coverage of Geographic Data in nuScenes}. {\color{PineGreen}Green} denotes frames with reliable geographic data;  {\color{BurntOrange}Orange} denotes samples where geographic data were retrieved but deemed unreliable and  excluded from dataset;
{\color{Mahogany}Red} denotes samples lacking any geographic data.}\vspace{-3mm}
    \label{fig:dataset-splits} 
\end{figure*}

\noindent\textbf{Equirectangular Panorama Representation for Efficient Storage and Retrieval.} Since the spatial sampling frequency of street view images is significantly lower than the key frame rate of nuScenes, multiple frames may correspond to the same street view location, as in Fig.~\ref{fig:dataset-matching}. To minimize storage overhead, each unique geographic data is retrieved only once and we store a mapping between each geographic data and all its nearest nuScenes frames.

However, different frames in nuScenes require different perspectives for the same street view location. To ensure geometrically correct views while maintaining storage efficiency, we adopt equirectangular panoramic representation: 
\begin{enumerate}
    \item \textbf{Data Fetching}: For each street view location, we retrieve 18 perspective images from API with distributed yaw angles (covering $360^\circ$) at a fixed pitch of $0^\circ$. 
    \item \textbf{Equirectangular Panorama Format}: These images are projected onto a spherical representation and stored in an equirectangular panoramic format.
    \item \textbf{Virtual Camera Alignment}:  For each nuScenes frame and each onboard camera, a virtual camera is instantiated at the corresponding street view location, with intrinsic parameters identical to the nuScenes camera model. The extrinsic transformation is derived from the ego pose and the street view capture point: the rotation follows the original nuScenes camera orientation, while the translation is computed from the latitude–longitude offset between the street view and the ego vehicle. The translation along the $z$-axis is set to a fixed constant.  
    \item \textbf{Reprojection Retrieval}: Using the virtual camera configuration, we perform perspective projection from the equirectangular panorama to synthesize a street view image, geometrically aligned with the nuScenes frame. 
\end{enumerate}
This process ensures spatial consistency and one-to-one correspondence between each onboard frame and its synthesized street view perspectives, while keeping the overall collection pipeline storage-efficient, reducing overall storage by more than 70\% compared to directly downloading per-frame street view crops.

\noindent\textbf{Dealing with Missing and Misaligned Geographic Data.} As shown in Fig.~\ref{fig:failure_case}, the Google Maps APIs might return an empty response or misaligned geographic data. As discussed in Sec.~\ref{sec:adaptive_fusion}, we design an adaptive fusion mechanism to let the model be able to selectively fuse reliable geographic information by residual gating. During the curation of nuScenes-Geography, we manually check all the downloaded geographic images and identify 1800 misaligned cases, serving as the negative labels of the Reliability Estimation module. Fig.~\ref{fig:dataset-splits} gives an overview of the coverage of geographic data for nuScenes, which is relatively high.

\begin{table*}[ht!]
\centering
\setlength{\tabcolsep}{17pt}     
\caption{\textbf{Online Mapping Results.} All methods use the ResNet50 backbone and are reproduced. Adding geographic priors brings significant mAP improvements to baselines. }
\vspace{-2mm}
\resizebox{\linewidth}{!}{
\begin{tabular}{lc|cccc}
\toprule
Method      & Epoch & AP$_{ped}$($\uparrow$)& AP$_{div}$($\uparrow$)& AP$_{bound}$($\uparrow$) & mAP($\uparrow$) \\ \midrule
MapTR       & 24    & 46.3                           & 51.5                           & 53.1                           & 50.3                           \\
\rowcolor{gray!10}
MapTR+Geo   & 24    & 56.5 (\textcolor{blue}{+10.2}) & 67.3 (\textcolor{blue}{+15.8}) & 59.8 (\textcolor{blue}{+6.7})  & 61.2 (\textcolor{blue}{+10.9}) \\
MapTR       & 110   & 55.9                           & 60.9                           & 61.1                           & 59.3                           \\
\rowcolor{gray!10}
MapTR+Geo   & 110   & 72.5 (\textcolor{blue}{+16.6}) & 75.0 (\textcolor{blue}{+14.1}) & 70.5 (\textcolor{blue}{+9.4})  & 72.7 (\textcolor{blue}{+13.4}) \\
MapTRv2     & 24    & 59.8                           & 62.4                           & 62.4                           & 61.5                           \\
\rowcolor{gray!10}
MapTRv2+Geo & 24    & 74.4 (\textcolor{blue}{+14.6}) & 74.7 (\textcolor{blue}{+12.3}) & 71.1 (\textcolor{blue}{+8.7})  & 73.4 (\textcolor{blue}{+11.9}) \\
MapTRv2     & 110   & 68.1                           & 68.3                           & 69.7                           & 68.7                            \\
\rowcolor{gray!10}
MapTRv2+Geo & 110   & 79.8 (\textcolor{blue}{+11.7}) & 77.4 (\textcolor{blue}{+9.1})  & 77.3 (\textcolor{blue}{+7.6})  & 78.2 (\textcolor{blue}{+9.5}) \\ 
\bottomrule
\end{tabular}}
\vspace{-2mm}
\label{tab:mapping_geo}
\end{table*}

\begin{table*}[t!]
\centering
\setlength{\tabcolsep}{5pt} 
\small 
\caption{\textbf{Occupancy Results on Occ3D-nuScenes.} All methods are reproduced.  Geographic priors improve mIoU especially for static terrain, considering it introduces extra information about background.\label{tab:occupancy_merged}}
\vspace{-2mm}
\begin{tabular}{l c | c c c c c c}
\toprule
\multirow{2}{*}{Method} & \multirow{2}{*}{Overall mIoU ($\uparrow$)} & \multicolumn{6}{c}{Per-Class mIoU ($\uparrow$) (Static Terrain)} \\
\cmidrule(lr){3-8}
& & driveable & other flat & sidewalk & terrain & manmade & vegetation \\
\midrule
FBOcc           & 39.11 & 80.07 & 42.76 & 51.18 & 55.13 & 42.19 & 37.53 \\
\rowcolor{gray!10}
FBOcc+Geo       & 39.74 (\textcolor{blue}{+0.63}) & 82.47 (\textcolor{blue}{+2.4}) & 44.47 (\textcolor{blue}{+1.71}) & 53.26 (\textcolor{blue}{+2.08}) & 57.7 (\textcolor{blue}{+2.57}) & 42.61 (\textcolor{blue}{+0.42}) & 38.57 (\textcolor{blue}{+1.04}) \\
\midrule
FlashOCC:M1     & 31.92 & 77.95 & 36.61 & 47.17 & 50.92 & 36.39 & 31.08 \\
\rowcolor{gray!10}
FlashOCC:M1+Geo & 32.35 (\textcolor{blue}{+0.43}) & 78.72 (\textcolor{blue}{+0.77}) & 39.51 (\textcolor{blue}{+2.9}) & 47.88 (\textcolor{blue}{+0.71}) & 51.82 (\textcolor{blue}{+0.9}) & 37.28 (\textcolor{blue}{+0.89}) & 31.94 (\textcolor{blue}{+0.86}) \\
\bottomrule
\end{tabular}
\vspace{-4mm}
\end{table*}

\begin{table}[ht!]
\centering
\setlength{\tabcolsep}{4pt}
\caption{\textbf{Object Detection Results.} The influence is marginal since geographic data mainly helps background.}
\vspace{-2mm}
\resizebox{\linewidth}{!}{
\begin{tabular}{l|c|cc}
\toprule
Method & Backbone & NDS ($\uparrow$) & mAP ($\uparrow$) \\
\midrule
BEVDet        & R50      & 39.41 & 30.85 \\
\rowcolor{gray!10}
BEVDet+Geo    & R50      & 39.43(\textcolor{blue}{+0.02}) & 30.69(\textcolor{red}{-0.16}) \\
BEVFormer     & R101-DCN & 51.70 & 41.60 \\
\rowcolor{gray!10}
BEVFormer+Geo & R101-DCN & 51.80(\textcolor{blue}{+0.10}) & 41.64(\textcolor{blue}{+0.04}) \\
\bottomrule
\end{tabular}}
\vspace{-3mm}
\label{tab:detection_geo}
\end{table}

\section{Experiments}
\label{sec:experiments}
In this section, we evaluate the proposed spatial retrieval paradigm on the extended nuScenes-Geography dataset across five tasks. We study three potential advantages of adopting spatial retrieval: enhancing static scene understanding, improving planning robustness, and enhancing spatial consistency of generative world model.

\subsection{Scene Understanding}

Spatial retrieval provides a stable view of the background, compensating for the vulnerability to extreme visual conditions and limited perception horizon of onboard sensors. 

\noindent\textbf{Online Mapping} As in Table~\ref{tab:mapping_geo}, integrating geographic priors into MapTR~\cite{liao2022maptr} and MapTRv2~\cite{liao2025maptrv2} substantially improves online mapping. The extra background information enables recovery of occluded lanes, as in Fig.~\ref{fig:static_scene_vis}.

\noindent\textbf{Occupancy Prediction} As in Table~\ref{tab:occupancy_merged}, Extending FB-OCC~\cite{li2023fb} and FlashOCC~\cite{yu2023flashocc} shows consistent mIoU gains, especially on static categories. The prior anchors background geometry against sensor noise, as in Fig.~\ref{fig:static_scene_vis}.

\noindent\textbf{Object Detection} As in Table~\ref{tab:detection_geo},  BEVDet~\cite{huang2021bevdet} and BEVFormer~\cite{li2022bevformer} show a negligible improvement with geographic data, which is natural since spatial retrieval mainly brings background information. However, adopting geographic data to distinguish foreground and background and then help object detection is an interesting future direction.


\subsection{Planning Robustness}

We assess how spatial retrieval facilitates safer planning with VAD~\cite{jiang2023vad}. The geographic prior provides consistent road layout information, compensating for sensing instability caused by occlusions or poor lighting. As shown in Table~\ref{tab:planning_geo}, while maintaining comparable trajectory accuracy, our method improves safety margins. Specifically, in challenging night scenes, it reduces the average collision rate from 0.55\% to 0.48\%, demonstrating the value of geographic priors as a robust guide for safe planning. Fig.~\ref{fig:planning_vis} visualizes examples.

\begin{table}[htpb!]
\centering
\setlength{\tabcolsep}{6pt}
\caption{\textbf{Generative World Model Results.} The empirical results validate the effectiveness of geographic priors. }
\vspace{-2mm}
\resizebox{\linewidth}{!}{
\begin{tabular}{l|c|cc}
\toprule
Method & Frame Size & FVD ($\downarrow$) & FID ($\downarrow$) \\
\midrule
 UVG& 35$\times$256$\times$448 & 36.10 & 5.82 \\
 \rowcolor{gray!10} 
 UVG+Geo& 35$\times$256$\times$448 & 29.97(\textcolor{blue}{+6.04}) & 5.60(\textcolor{blue}{+0.22}) \\
 MDD& 17$\times$224$\times$400 & 84.43 & 18.38 \\
 \rowcolor{gray!10} 
 MDD+Geo& 17$\times$224$\times$400 & 81.52(\textcolor{blue}{+2.91}) & 18.10(\textcolor{blue}{+0.28}) \\
\bottomrule
\end{tabular}}
\vspace{-3.5mm}
\label{tab:gen_geo}
\end{table}

\begin{figure}[t!]
    \centering
    \includegraphics[width=1.0\columnwidth]{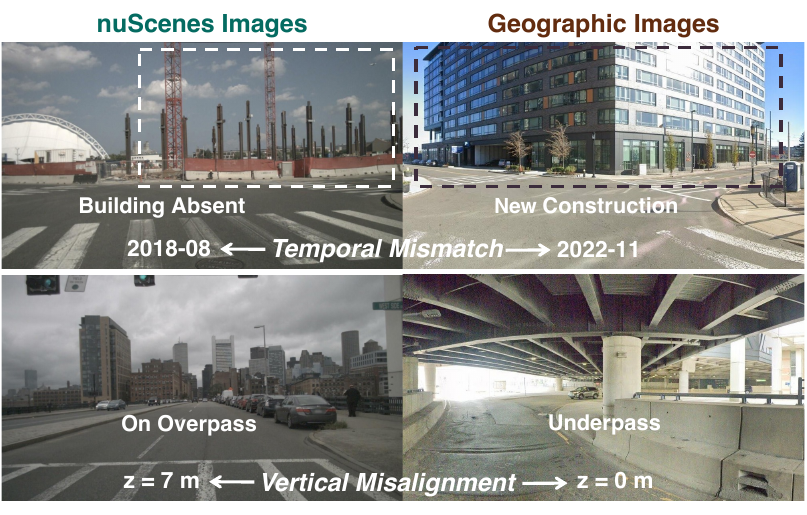}
    \vspace{-6mm}
    \caption{\textbf{Visualization of Misaligned Spatial Retrieval.} 
    Examples include a \textbf{temporal mismatch} (top) due to scene changes and a \textbf{vertical misalignment} (bottom) from altitude errors. Such discrepancies can lead to incorrect maps or occupancy predictions.}
    \vspace{-6mm}
    \label{fig:failure_case}
\end{figure}

\begin{figure}[t!]
    \centering
    \includegraphics[width=0.98\columnwidth]{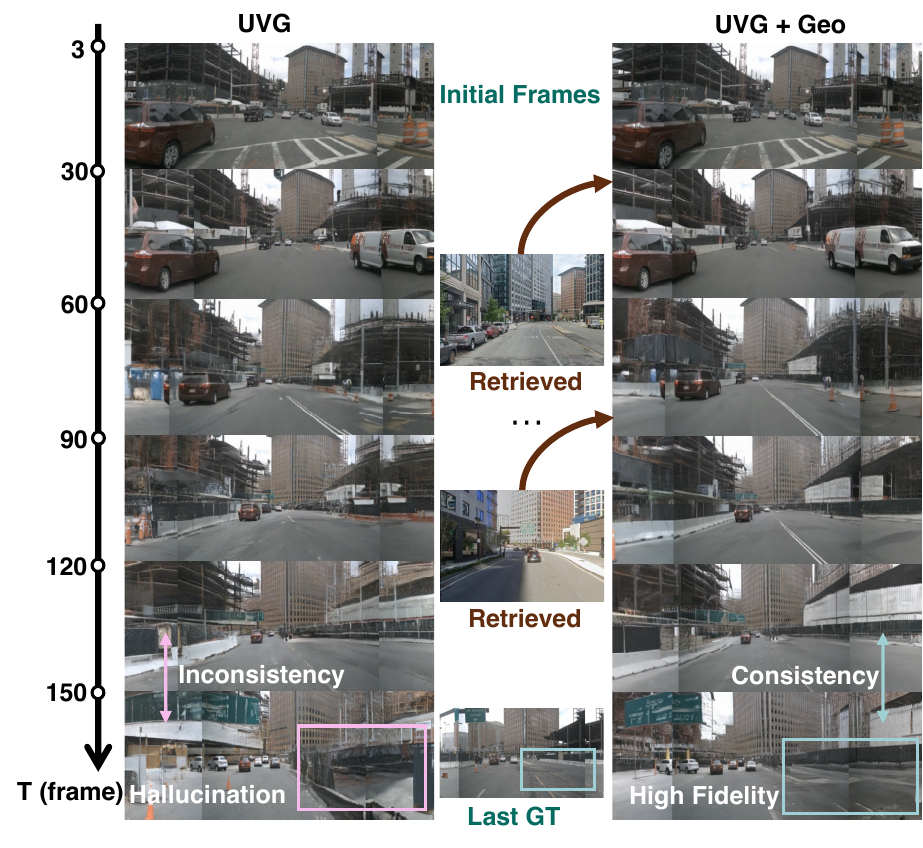} 
    \vspace{-2mm}
    \caption{\textbf{Generative World Model Case Study.} Geographic data provides spatial clues, enhancing temporal consistency and mitigating hallucinations.
    }
    \vspace{-2mm}
    \label{fig:quality_gen} 
\end{figure}

\begin{table*}[ht!]
\centering
\setlength{\tabcolsep}{10pt}
\caption{\textbf{End-to-end Planning Results.} 
All methods are reproduced. Introducing geographic information achieves comparable planning accuracy in terms of L2 error. On the challenging night subset, VAD+Geo has lower collision rates, indicating improved safety performance.}
\vspace{-2mm}
\resizebox{1.0\textwidth}{!}{
\begin{tabular}{l|l|cccc|cccc}
\toprule
\multirow{2}{*}{Method} & \multirow{2}{*}{Split} &
\multicolumn{4}{c|}{L2 (m) $\downarrow$} &
\multicolumn{4}{c}{Collision (\%) $\downarrow$} \\
& & 1s & 2s & 3s & Avg. & 1s & 2s & 3s & Avg. \\
\midrule
VAD & Full & \textbf{0.34} & \textbf{0.61} & 0.98 & \textbf{0.64} & 0.16 & \textbf{0.29} & 0.66 & 0.37 \\
\rowcolor{gray!10}
VAD+Geo & Full & 
0.35(\textcolor{red}{-0.01}) & 
0.62(\textcolor{red}{-0.01}) &
\textbf{0.96}(\textcolor{blue}{+0.02}) & 
\textbf{0.64}(\textcolor{gray}{0.00}) &
\textbf{0.14}(\textcolor{blue}{+0.02}) & 
\textbf{0.29}(\textcolor{gray}{0.00}) &
\textbf{0.64}(\textcolor{blue}{+0.02}) & 
\textbf{0.36}(\textcolor{blue}{+0.01}) \\
\midrule
VAD & Night & \textbf{0.44} & \textbf{0.82} & 1.35 & \textbf{0.87} & 0.10 & 0.24 & \textbf{1.30} & 0.55 \\
\rowcolor{gray!10}
VAD+Geo & Night & 
0.46(\textcolor{red}{-0.02}) & 
0.83(\textcolor{red}{-0.01}) &
\textbf{1.33}(\textcolor{blue}{+0.02}) & 
\textbf{0.87}(\textcolor{gray}{0.00}) &
\textbf{0.00}(\textcolor{blue}{+0.10}) & 
\textbf{0.15}(\textcolor{blue}{+0.09}) &
\textbf{1.30}(\textcolor{gray}{0.00}) & 
\textbf{0.48}(\textcolor{blue}{+0.07}) \\
\bottomrule
\end{tabular}}
\vspace{-1mm}
\label{tab:planning_geo}
\end{table*}

\begin{table*}[!tb] 
\centering
\caption{\textbf{Ablation Study} with Occupancy Task (Static mIoU for non-movable
object categories) and Generative World Model Task (FVD).\label{tab:ablation}}
\vspace{-2mm}
\subcaptionbox{\textbf{Effect of Geo Images}\label{tab:geo}}{
    \begin{adjustbox}{width=0.27\textwidth} 
    \setlength{\tabcolsep}{5pt}
    \begin{tabular}{l|c|c}
    \toprule
    Variant &  S mIoU ↑ & FVD ↓ \\ \midrule
    w/o Geo Images    & 46.66 & 35.42 \\
    \rowcolor{gray!10}
    w Geo Images      & 47.86 & 29.97 \\
    \bottomrule
    \end{tabular}
    \end{adjustbox}
}
\hfill 
\subcaptionbox{\textbf{Effect of Positional Encoding}\label{tab:3dpe}}{
    \begin{adjustbox}{width=0.23\textwidth} 
    \setlength{\tabcolsep}{5pt}
    \begin{tabular}{l|c|c}
    \toprule
    Variant     & S mIoU ↑ & FVD ↓ \\ \midrule
    w/o 3DPE    & 46.22  & 32.82 \\
    \rowcolor{gray!10}
    w 3DPE      & 47.86  & 29.97 \\
    \bottomrule
    \end{tabular}
    \end{adjustbox}
}
\hfill 
\subcaptionbox{\textbf{Effect of Reliability Estimation Gate(REG)}\label{tab:ref_gate}}{
    \begin{adjustbox}{width=0.35\textwidth} 
    \setlength{\tabcolsep}{17pt}
    \begin{tabular}{l|c|c}
    \toprule
    Variant  & S mIoU ↑ & FVD ↓ \\ \midrule
    w/o REG  & 47.65  & 30.95 \\
    \rowcolor{gray!10}
    w REG    & 47.86  & 29.97 \\
    \bottomrule
    \end{tabular}
    \end{adjustbox}
}
\vspace{-3mm} 
\end{table*}

\begin{figure}[!t]
    \centering
    \includegraphics[width=0.9\columnwidth]{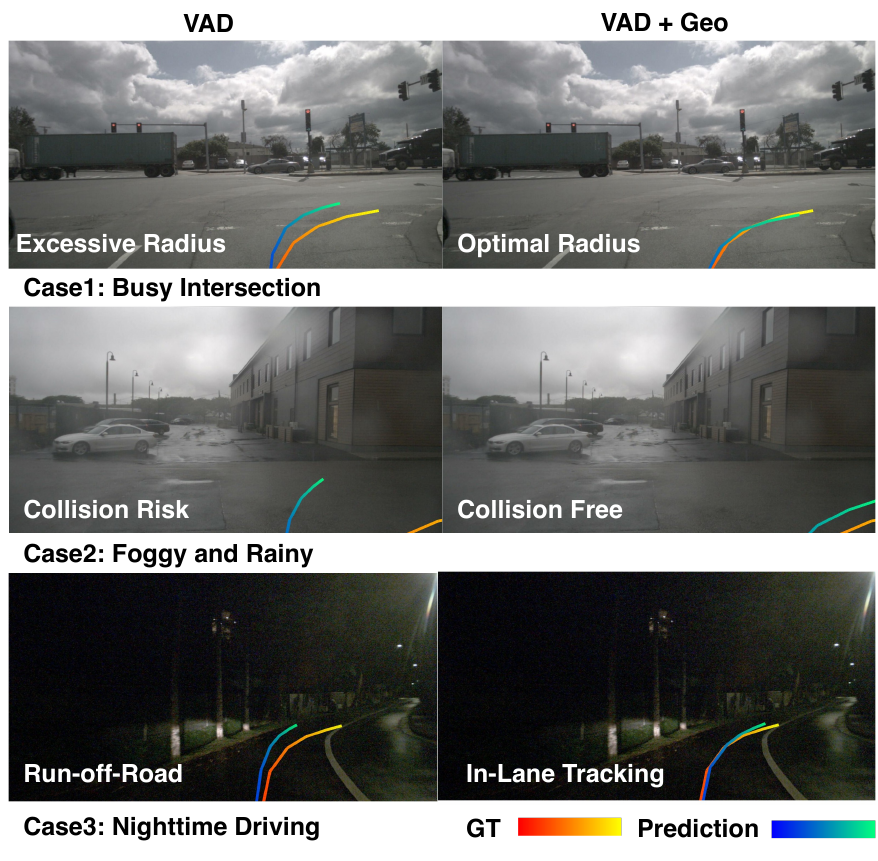} 
    \caption{\textbf{End-to-End Planning Case Study.} Geographic priors lead to better trajectories in complex intersections, night conditions, and adverse weather.}
    \label{fig:planning_vis}
    \vspace{-4mm}
\end{figure}

\subsection{Generative World Model Consistency}

We further evaluate whether geographic priors help generative world models.  
Conditioning UniMLVG~\cite{unimlvg} and MagicDriveDit~\cite{gao2024magicdrivedit} (For MagicDriveDit, we adjust the test-set sampling stride to 13 to avoid repeatedly sampling near-duplicate segments.) on geographic images yields lower FVD and FID, preventing scene drift and maintaining geometric consistency during rollouts, as in Table~\ref{tab:gen_geo}. This confirms that spatial retrieval acts as a structural scaffold for coherent world modeling.

\subsection{Visualization of Misaligned Spatial Retrieval}

The new paradigm faces the challenge of missing or misaligned retrieval issues, which occur when the offline geographic images are inconsistent with the camera images, as in Fig.~\ref{fig:failure_case}. This can occasionally happen due to: \textbf{(1) Outdated Maps:} Road layouts change due to construction, but the cached map imagery does not accurately reflect this, potentially confusing the model. \textbf{(2) GPS/Localization Error:} Inaccurate ego-pose can lead to misalignment between retrieved images and  onboard sensor images, which sometimes happen with Google Maps APIs.

\subsection{Ablation Study}
We conduct ablation studies with occupancy task (FlashOcc~\cite{yu2023flashocc}) and generative world model task (Unimlvg~\cite{unimlvg}).
As in Table~\ref{tab:ablation}, we observe that introducing geographic priors always brings explicit performance gains while positional encoding and reliability ability estimation gate bring further gains.

\subsection{Qualitative Analysis}
\begin{itemize}
    \item \textbf{Online Mapping}: Fig.~\ref{fig:static_scene_vis} shows the geographic priors help reconstruct map elements when visual cues from the onboard camera are degraded or missing.
    \item \textbf{Occupancy Prediction}: Fig.~\ref{fig:static_scene_vis} shows the geographic priors provide a clean and stable geometric reference, enabling the recovery of obscured background structures.
    \item \textbf{Planning}: Fig.~\ref{fig:planning_vis} shows that the stable road geometry from the geographic prior enables smoother and safer trajectories in complex intersections and adverse weather.
    \item \textbf{Generative World Modeling}: Fig.~\ref{fig:quality_gen} shows that the geographic priors 
    prevent generative collapse during long-horizon rollouts, maintaining scene consistency.
\end{itemize}

\begin{figure*}[t!]
    \centering
    \includegraphics[width=0.92\textwidth]{figs/static_w_anno_2.pdf} 
    \vspace{-3mm}
    \caption{\textbf{Geographic Priors Enhance Scene Understanding in Challenging Scenarios.} Our method leverages geographic priors to correct errors in online mapping (top) and occupancy prediction (bottom) under challenging conditions.}
    \label{fig:static_scene_vis}
    \vspace{-3mm}
\end{figure*}

\subsection{Robustness to Inaccurate Retrieval}
To further evaluate the effectiveness of the proposed Reliability Estimation Gate, we assess online mapping method - MapTRv2's~\cite{liao2025maptrv2} robustness under inaccurate retrieval.

\begin{figure}[t!]
    \centering
    \includegraphics[width=0.9\columnwidth]{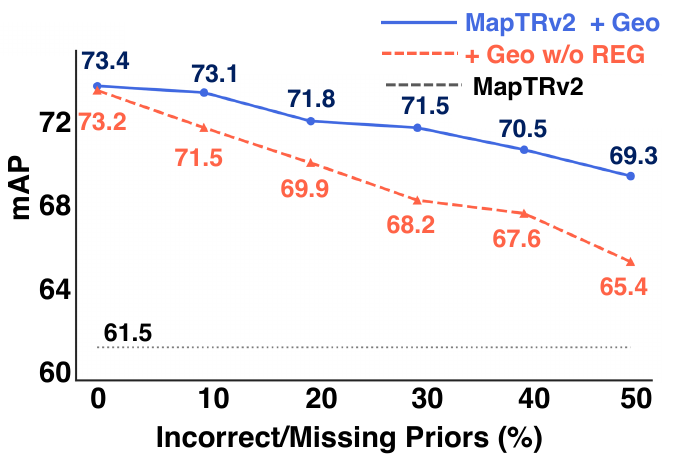}
    \vspace{-2mm}
    \caption{\textbf{Robustness to Inaccurate Retrieval.} Performance of MapTRv2+Geo as a function of the percentage of frames where the geographic images are drop or replaced with wrong images. Performance degrades gracefully, demonstrating the model is not overly reliant on the prior. Notably, without the Reliability Estimation Gate (REG), the performance decline is steeper, indicating reduced robustness to inaccurate priors.}
    \label{fig:robustness_analysis}
    \vspace{-8mm}
\end{figure}

We randomly drop the geographic images or replace them with random incorrect images for a certain percentage of frames. Figure~\ref{fig:robustness_analysis} shows that the model's performance degrades gracefully as the availability of the prior decreases. Even with 50\% of the priors missing or misaligned, the model retains a significant portion of its performance gain over the baseline. This indicates that the proposed Reliability
Estimation Gate allows the model to leverage the prior when available but does not lead to catastrophic failure in its absence, demonstrating robust real-world applicability.
\vspace{10mm}

\section{Conclusion}
In this work, we present the spatial retrieval paradigm for AD, introducing geographic data as an additional input. We extend nuScenes with geographic data by Google Maps APIs and evaluate five key AD tasks on the extended nuScenes-Geography dataset. We propose a general plug-and-play Spatial Retrieval Adapter module as an intuitive baseline to incorporate geographic data. We propose Reliability Estimation  to adaptively fuse geographic information based on the reliability of the retrieved data. Extensive experiments show that the proposed paradigm can enhance the performance of multiple AD tasks, demonstrating the substantial potential of the new paradigm.

\section{Limitation \& Future Work}To study the proposed spatial retrieval paradigm, we adopt Google Maps APIs to obtain geographic data. In practical usage, AD companies could use their huge offline traverse dataset of deployed cities, which might have higher coverage and image quality. For simplicity and generality, we only retrieve the nearest geographic image and only use the cross attention for fusion. Future work could explore more advanced, task-specific retrieval strategies and neural network designs for further performance enhancement.

\clearpage

\makeatletter
\providecommand{\maketitle}{}%
\providecommand{\bibliography}[1]{}%
\providecommand{\bibliographystyle}[1]{}%
\providecommand{\tableofcontents}{}%
\providecommand{\section}{}%
\providecommand{\subsection}{}%
\providecommand{\subsubsection}{}%

\let\cvpr@origmaketitle\maketitle
\def\maketitle{}%

\let\cvpr@origbibliography\bibliography
\let\cvpr@origbibliographystyle\bibliographystyle
\def\bibliography#1{}%
\def\bibliographystyle#1{}%

\let\cvpr@origtoc\tableofcontents
\def\tableofcontents{\@starttoc{suptoc}}%

\let\cvpr@origaddcontentsline\addcontentsline
\newcommand{\cvpr@supp@writetoc}[2]{%
  \begingroup
  \edef\cvpr@tocentry{\detokenize{#2}}%
  \cvpr@origaddcontentsline{suptoc}{#1}{\cvpr@tocentry}%
  \endgroup
}%

\let\cvpr@origsection\section
\renewcommand{\section}{\@ifstar\cvpr@supp@sectionstar\cvpr@supp@section}
\newcommand{\cvpr@supp@section}[1]{%
  \cvpr@origsection{#1}%
  \cvpr@supp@writetoc{section}{#1}%
}
\newcommand{\cvpr@supp@sectionstar}[1]{%
  \cvpr@origsection*{#1}%
  \cvpr@supp@writetoc{section}{#1}%
}

\let\cvpr@origsubsection\subsection
\renewcommand{\subsection}{\@ifstar\cvpr@supp@subsectionstar\cvpr@supp@subsection}
\newcommand{\cvpr@supp@subsection}[1]{%
  \cvpr@origsubsection{#1}%
  \cvpr@supp@writetoc{subsection}{#1}%
}
\newcommand{\cvpr@supp@subsectionstar}[1]{%
  \cvpr@origsubsection*{#1}%
  \cvpr@supp@writetoc{subsection}{#1}%
}

\let\cvpr@origsubsubsection\subsubsection
\renewcommand{\subsubsection}{\@ifstar\cvpr@supp@subsubsectionstar\cvpr@supp@subsubsection}
\newcommand{\cvpr@supp@subsubsection}[1]{%
  \cvpr@origsubsubsection{#1}%
  \cvpr@supp@writetoc{subsubsection}{#1}%
}
\newcommand{\cvpr@supp@subsubsectionstar}[1]{%
  \cvpr@origsubsubsection*{#1}%
  \cvpr@supp@writetoc{subsubsection}{#1}%
}
\makeatother

\makeatletter
\let\cvpr@origat\@
\def\@{\ifhmode\spacefactor\@m\else\relax\fi}%
\makeatother


\maketitle

\appendix
\onecolumn

\section*{Supplementary Material}
\label{sec:supplementary}

\tableofcontents
\clearpage


\section{Discussion}

\subsection{How does the latency of spatial retrieval influence the system?}
In our experimental pipeline, all geographic images are retrieved, aligned, and processed \emph{offline} before training and evaluation. This ensures that, during inference, the stack requires no interaction with external services (i.e. Google Maps APIs). 

For real-world autonomous driving deployments, the geographic data can similarly be pre-downloaded and cached on the vehicle or on edge servers, as the storage cost is low due to the equirectangular panoramic representation that stores each street view location only once. Considering the required storage footprint is modest, the data can be bulk-downloaded during periods of reliable connectivity, allowing the inference stage to operate without relying on real-time network access and without being affected by bandwidth fluctuations.

Importantly, as demonstrated in our experiments, in cases where geographic images are unavailable during driving—such as regions with insufficient coverage or operation in offline-only settings—the Reliability Estimation Gate automatically down-weights or suppresses the geographic branch. The system thus reverts to the baseline perception pipeline without compromising safety or prediction quality, maintaining a stable and predictable lower-bound performance. Overall, the combination of offline acquisition, compact storage design, and modest computation makes the paradigm suitable for realistic deployment scenarios while avoiding latency overhead. 

\subsection{How does the method behave when the spatial retrieval data are missing, outdated, or unreliable?}
Geographic images are inherently heterogeneous in density and recency, and discrepancies may arise due to construction, occlusions at capture time, or environmental changes. The proposed framework explicitly accounts for such imperfect conditions through the Reliability Estimation Gate, which evaluates both spatial distance and visual similarity between onboard and retrieved features. When retrieved images contain mismatched content, outdated structures, or notable photometric differences, the gate attenuates their contribution, preventing unreliable geographic cues from degrading the prediction.

The degradation experiment (Fig. 11 in the main paper) validates the robustness of the system by simulating missing or incorrect retrieval data. As the proportion of unavailable or perturbed data increases, performance declines smoothly and without catastrophic failure, demonstrating that those tasks remain supported by onboard-sensor alone. In other words, the model does not become overly dependent on geographic information and incorporates it only when reliable. This balanced fusion behavior ensures that the paradigm remains stable across a wide range of real-world conditions, including sparse coverage, partial misalignment, or outdated images.




\subsection{Why do some tasks benefit more moderately from geographic context?}

The geographic modality predominantly captures static environmental structure—road geometry, lane topology, sidewalks, and background scene layout—while lacking dynamic elements such as moving vehicles or pedestrians. Tasks that rely on static spatial organization, such as online mapping or occupancy prediction, naturally gain more from this prior. Improvements arise from the ability of geographic images to provide visibility into regions that may be occluded, poorly lit, or beyond the instantaneous perception horizon of onboard sensors. 

For tasks whose performance is dominated by dynamic-object interpretation—such as object detection—the contribution of geographic priors is inherently more limited. The spatial retrieval paradigm is neither intended nor designed to replace online sensor information in such settings, but rather to serve as a reliable auxiliary cue. The more moderate gains observed for these tasks align with the role of static geographic information as an additive structural prior rather than a dynamic-sensing modality. This task-dependent pattern is consistent with the modality’s characteristics and supports the intended complementary nature of spatial retrieval.

\subsection{Is the proposed paradigm restricted to Google Maps as the source of geographic images?}

Although Google Maps provides a convenient and accessible source of geographically anchored images for our experiments, the proposed paradigm is agnostic to the data provider. The method requires only panoramic or perspective images accompanied by approximate capture coordinates. Thus, it can be directly applied to images acquired through alternative mapping platforms or to large-scale traversal logs collected by autonomous driving fleets.

In practical deployments, companies may employ their own high-resolution street-level datasets, which often feature denser coverage and more consistent capture conditions than public sources. Because the retrieval function, panoramic representation, and fusion module rely only on pose and appearance alignment, no architectural modifications are required to accommodate different geographic data sources. This highlights that the core contribution lies in the spatial retrieval paradigm itself, rather than characteristics of any particular geographic images provider.

\section{More Related Work}

\subsection{Camera-based 3D Detection}

Bird's-eye-view (BEV) object detection~\cite{vpn,lss,bevdet,detr3d,petr,simmod,bevformer,bevformerv2,solofusion} aims to detect objects in BEV space from single-view or multi-view 2D images. Early works~\cite{oft,pseudo-lidar,caddn} transform 2D features into BEV based on single-view images for monocular 3D detection. LSS~\cite{lss} extends this to multi-view input and lifts image features into 3D via depth estimation. Following LSS, BEVDet~\cite{bevdet} lifts 2D features to BEV and applies a BEV encoder with residual blocks and FPN, while BEVDepth~\cite{bevdepth} shows that explicit depth supervision and efficient BEV Pooling~\cite{bevdet,bevdepth,bevpoolv2} are crucial for both accuracy and efficiency in the 2D-to-3D lifting pipeline. In contrast, BEVFormer~\cite{bevformer} introduces a spatio-temporal Transformer encoder that directly constructs BEV representations from multi-view, multi-timestamp inputs via deformable cross-attention, and BEVFormerV2~\cite{bevformerv2} further eases optimization with perspective-view supervision and joint monocular/BEV supervision. From the view transformation perspective, BEVFormer and its variants~\cite{bevformer,bevformerv2,heightformer} adopt dense BEV queries with heavy deformable attention, whereas BEVDet/BEVDepth~\cite{bevdet,bevdepth} follow an LSS-style~\cite{lss} 2D-to-3D lifting plus BEV pooling pipeline. Inspired by DETR, sparse query-based BEV methods~\cite{detr3d,petr,petrv2} instead start from a small set of 3D queries and interact with image features either via 3D-to-2D projection~\cite{detr3d} or global attention with 3D positional embeddings~\cite{petr,petrv2}, trading dense BEV grids for a compact query set but often at the cost of expensive global attention and limited scalability to long-term temporal fusion.

\subsection{Online Mapping}

Online mapping models~\cite{hdmapnet,vectormapnet,Streammapnet,maptr,SemVecNet} tackle autonomous driving by dynamically generating map geometry and semantics from sensor inputs, thereby reducing reliance on manual annotations and enabling adaptive mapping in changing environments. With the development of PV-to-BEV methods~\cite{Ma2022VisionCentricBP,li2022delving}, HD map construction is often formulated as BEV segmentation on surround-view images, where many approaches~\cite{gkt,cvt,fiery,bevformer,lss,polarbev,liu2022bevfusion,baeformer,uniformer,beverse} produce rasterized maps through BEV semantic segmentation. To obtain vectorized HD maps, HDMapNet~\cite{hdmapnet} performs heuristic and time-consuming post-processing on pixel-wise segmentation, while VectorMapNet~\cite{vectormapnet} is the first end-to-end method with a two-stage coarse-to-fine design and an autoregressive decoder, suffering from long inference time and permutation ambiguity. MapTR~\cite{maptr} and StreamMapNet~\cite{Streammapnet} further leverage monocular or multi-view images for end-to-end online HD map construction, enhancing deployment flexibility, but most existing models are trained on fixed datasets and tightly coupled to specific sensor setups and environments, leading to overfitting and degraded performance under domain shift. SemVecNet~\cite{SemVecNet} mitigates cross-dataset variation via an intermediate semantic map, yet still falls short of the robustness required for real-world deployment. In contrast, our method introduces a unified shape modeling strategy for map elements that resolves permutation ambiguity and stabilizes training, and further builds a structured, parallel one-stage framework with much higher efficiency. Concurrent and follow-up works of our conference version~\cite{maptr} explore alternative end-to-end HD map constructions~\cite{shin2023instagram,qiao2023end,zhang2023online,chen2023polydiffuse,qiao2023machmap,li2023topology,yuan2024streammapnet,xu2023insightmapper,li2024lanesegnet,yuan2024presight,chen2024maptracker,jiang2024p,wang2024stream,xiong2024ean,kalfaoglu2023topomask,liu2024leveraging} and extend to related tasks~\cite{nmp,toponet,lanegap,chen2023vma,pip,vad}, including Bézier- and pivot-based map representations~\cite{qiao2023end,pivotnet}, neural global map representations~\cite{nmp}, topology-aware modeling~\cite{toponet}, diffusion-based refinement~\cite{chen2023polydiffuse}, differentiable rasterization~\cite{zhang2023online}, and generalized centerline and 3D map learning in MapTRv2.

\subsection{Occupancy Prediction}
Occupancy prediction has recently emerged as a 3D alternative to conventional bird’s-eye view (BEV) perception. BEV representations~\cite{li2023delving,ma2022vision} provide an occlusion-reduced and metrically consistent top-down view that is well suited for multi-view, multimodal, and temporal fusion, as well as downstream planning and decision making. However, BEV inevitably collapses the height dimension and thus cannot fully capture fine-grained 3D geometry or vertical structures. Occupancy perception addresses this limitation by estimating the occupied state of voxels in a discretized 3D space. Such dense 3D fields naturally support open-set objects, irregular shapes, complex road structures, and occlusion reasoning~\cite{occupancy_network_blog,wang2023panoocc}, and can be jointly used for 3D detection, segmentation, and tracking~\cite{peng2023learning,zhou2024sogdet,wang2023panoocc,min2024driveworld}. Recent work further enriches occupancy with semantics~\cite{openocc}, language priors~\cite{vobecky2024pop}, and motion cues~\cite{cam4docc}, making vision-centric occupancy a cost-effective alternative to LiDAR-heavy systems.

In the camera-only setting, most methods still build on BEV-style formulations to infer voxel-wise occupancy from multi-view images. One line of work directly learns voxel features in 3D space: VoxFormer~\cite{li2023voxformer} uses 2.5D cues to initialize voxel queries, Occ3D~\cite{tian2023occ3d} adopts a coarse-to-fine voxel encoder, and RenderOcc~\cite{pan2023renderocc} predicts density and semantics under NeRF-style supervision, facilitated by dense occupancy benchmarks~\cite{sima2023_occnet,tian2023occ3d}. Another line couples occupancy prediction more tightly with BEV representations. TPVFormer~\cite{huang2023tri} decomposes the scene into three orthogonal views, SurroundOcc~\cite{wei2023surroundocc} lifts BEV features along the height axis with spatial cross-attention, OccNet~\cite{tong2023scene} builds a unified occupancy embedding tailored for planning, and FBOcc~\cite{li2023fb} designs a forward–backward view transformation scheme.

\subsection{End-to-End Planning}

Autonomous driving planning has evolved through several distinct stages, beginning with rule-based frameworks and gradually shifted toward learning-driven and end-to-end systems. Early planning modules were dominated by rule-based methods, where vehicle behavior followed manually designed rules and heuristic decision logic~\cite{Treiber_2000, fan2018baidu, Dauner2023CORL}. These planners offered strong interpretability and reliable control, and were successfully deployed in many real-world systems~\cite{Leonard2008APA, Urmson2008AutonomousDI}. Their fundamental limitation, however, lies in their inability to generalize: once the environment deviates from predefined scenarios, the decision-making process becomes brittle.

To improve adaptability, the community gradually shifted toward learning-based planners, framing planning as an imitation learning problem~\cite{uniad, tampuu2020survey, chen2023end}. Early behavior cloning approaches relied on CNNs~\cite{bojarski2016end, kendall2019learning, hawke2020urban} and RNNs~\cite{bansal2018chauffeurnet}, and were later extended to Transformer-based architectures~\cite{scheel2021urban, chitta2022transfuser} to better capture complex driving behaviors. While these models produce more human-like trajectories, imitation learning lacks multi-modality guarantees and is sensitive to distribution shift, resulting in compounding errors during closed-loop execution. Consequently, many systems still fall back on rule-based post-processing—either refining~\cite{vitelli2022safetynet, huang2023gameformer} or selecting~\cite{cheng2024pluto} trajectories—which runs counter to the original goal of eliminating handcrafted rules. Moreover, tuning behavior for safety or personalization is difficult, since training-time objectives often conflict with one another~\cite{bansal2018chauffeurnet, cheng2024pluto}.

Driven by the need for tighter integration between perception, prediction, and planning, planning has increasingly been studied under the umbrella of end-to-end autonomous driving (E2EAD). Modern end-to-end planners directly map sensor inputs to future trajectories or control actions~\cite{uniad, vad, vadv2, sparsedrive, P3, STP3, LAW, alphadrive, centaur, drivinggpt, drivetransformer, DonShakeWheel, recogdrive}, reducing the error accumulation inherent in multi-stage pipelines. UniAD~\cite{uniad} established a unified planning-oriented architecture; VAD~\cite{vad} and VADv2~\cite{vadv2} streamlined the representation and action space; SparseDrive~\cite{sparsedrive} leveraged sparse structures; and diffusion-based models such as DiffusionDrive~\cite{diffusiondrive} introduced generative planning via diffusion policies.

To further improve planning safety, recent efforts have explored score-based planning optimization. Hydra-MDP~\cite{hydra} distills both human demonstrations and rule-derived metrics through multiple scoring heads, while WOTE~\cite{wote} predicts future BEV states to evaluate trajectory quality. These approaches enhance safety supervision but still operate at the score level, independently optimizing each anchor without modeling the global policy distribution.

\subsection{Generative World Model}
Recent progress in driving video generation has advanced realistic and controllable autonomous driving simulation~\cite{gao2023magicdrive,gao2025vista,li2024hierarchical,gao2024magicdrivedit,li2023bridging,wang2024driving,mao2024dreamdrive,wang2024stag}. 
MagicDrive~\cite{gao2023magicdrive} achieves high-fidelity street-view synthesis through tailored encoders and cross-view attention for accurate 3D geometry control, while UniScene~\cite{li2024uniscene} unifies multi-modal data generation (semantic occupancy, video, LiDAR) via progressive processes. 
Extensions such as MagicDriveDiT~\cite{gao2024magicdrivedit} and MagicDrive3D~\cite{gao2024magicdrive3d} further improve and enhance scalability using DiT architectures and deformable Gaussian splatting for high-resolution 3D reconstruction.
DreamDrive~\cite{mao2024dreamdrive} synthesizes 4D spatiotemporal scenes via hybrid Gaussian representations, balancing visual fidelity and generalizability.

Beyond open-loop video synthesis, a related line of work explores using generative models as driving world models that predict future multi-view observations conditioned on controls and scene layouts. Unlike general-purpose video generation, generative world models for autonomous driving are designed to support multi-agent interaction, ego-motion control, environmental diversity, and coherent multi-camera observations. Early efforts in this direction include GAIA-1~\cite{gaia1-2023}, which introduces text and ego-action conditioning within a discrete world model equipped with a video diffusion decoder to enhance temporal consistency, and CommaVQ~\cite{comma2023vq}, which similarly employs a causal transformer over discrete tokens to model controllable ego-motion in driving scenes.Subsequent approaches build on latent diffusion models to achieve higher-fidelity generation and richer forms of control. DriveDreamer~\cite{wang2023drivedreamer} conditions a latent diffusion model on 3D bounding boxes, HD maps, and ego actions, and further adds an action decoder for future ego-action prediction. Drive-WM~\cite{wang2024driving} extends this paradigm to multi-camera settings with up to six surrounding views and introduces controllable ego behavior, dynamic agents, and environmental factors such as weather and lighting. UniMLVG~\cite{chen2024unimlvg} enables multi-view video synthesis conditioned on text, camera parameters, 3D bounding boxes, and HD maps, while MaskGWM~\cite{ni2025maskgwm} towards longer temporal horizons. Vista~\cite{gao2024vista} focuses on high-resolution, long-duration videos, and Delphi~\cite{ma2024unleashing} steers generation toward failure-prone scenarios to enhance the utility of synthetic data for training. 

\subsection{Autonomous Driving with Retrieval}

Recent research in autonomous driving has increasingly emphasized retrieval-augmented decision making, where external knowledge is queried and combined with on-board perception rather than relying solely on end-to-end models. Neuro-symbolic approaches \cite{Yi} and knowledge-graph-based methods \cite{Hogan, Martin} can be seen as early forms of retrieval, using symbolic structures to recall missing entities or relations (e.g., occluded pedestrians or lane-change intentions). In parallel, cooperative perception retrieves observations from other agents in the traffic network via V2X communication \cite{ren2022collaborative, han2023collaborative, yang2022autonomous, liu2023towards}, mitigating the “short-sightedness” of purely on-board perception \cite{miao2022does} and improving situation awareness, especially in occluded or dense scenes \cite{narri2021set, xiao2023overcoming, loh2024crossdomaintransferlearningusing}. Recent work such as CodeFilling \cite{sun2023toward, hu2024communication, 10.1007/978-3-031-71470-2_19} further moves from passive broadcast to more selective, representation-aware communication, which conceptually aligns with retrieval-style filtering.

More directly, a growing body of work treats driving scenarios as the core memory to be retrieved in Retrieval-Augmented Generation (RAG) pipelines. Structured scenario platforms (OpenScenario, MetaScenario, CommonRoad) \cite{openscenario, chang2023metascenario, althoff2017commonroad} and large-scale datasets \cite{chang2023metascenario, li2022features, hu2022scenario, zhang2022systematic, duan2024distributional, li2023onramp, hu2024noise, huang2024mfe, li2019integral, ge2024task, feng2021intelligent, chang2024vista} provide the basis for retrieving past interactions and corner cases. LLM-based Driving-RAG systems query similar scenarios and textual knowledge to support online planning and decision making \cite{hussien2024rag, yuan2024rag, xu2024drivegpt4, dai2024tiv} and offline scenario generation and simulation \cite{nguyen2024text, guo2024mixing, zhang2023cavsim}, while other works use LLMs with retrieved driving knowledge for instruction following and human–vehicle interaction \cite{cai2024driving, wang2023chatgpt, chang2024llm}. To make such retrieval effective, prior studies propose task-driven scenario embeddings \cite{malawade2022roadscene2vec, wang2024rs2g, wang2020clustering} and similarity metrics \cite{kerber2020clustering, hauer2020clustering, zipfl2023traffic, chang2024vista}, sometimes combined with distribution-aware indexing \cite{hu2022scenario, li2021foundation, malkov2020approximate} and graph-based post-processing \cite{wang2024rs2g, pan2024unifying} to refine the top-$K$ retrieved scenarios and reduce LLM confusion and hallucination.

\section{nuScenes-Geography Construction Details}

\subsection{Overview of the Construction Pipeline}

The objective of nuScenes-Geography extension is to augment each nuScenes frame with geographically grounded visual information retrieved from the Google Maps APIs, which include both the Google Street View API and the map‐tile interfaces.
Among these sources, street view images constitute the core component of our pipeline due to its need for viewpoint synthesis and geometric alignment; thus, it is the primary focus of the following sections.
In contrast, map data follow a significantly simpler procedure: we download the complete nuScenes regional map beforehand and crop the corresponding regions directly based on the transformed coordinates, requiring no additional viewpoint modeling.

The overall construction pipeline contains three major stages.
First, we convert nuScenes ego poses from their local map coordinate system into global geodetic coordinates (latitude–longitude), which enables precise querying of the Google Maps APIs.
Second, for each valid geographic location, we retrieve multi-view street view images via the Google Street View API and reconstruct an equirectangular panoramic representation that compactly encodes all viewing directions.
Finally, for every nuScenes camera frame, we instantiate a virtual camera at the matched street view location and synthesize a geometrically aligned street-level view through spherical-to-perspective reprojection.
This pipeline establishes a one-to-one correspondence between each onboard camera image and its aligned street view rendering, while maintaining high storage efficiency and broad geographic coverage across the dataset.

\subsection{Coordinate Systems and Geo-Localization}
\label{sec:supp_coordinates}

\paragraph{Ego-pose representation in nuScenes.}
We adopt the nuScenes global coordinate frame as the starting point for geo-localization.
For each driving scene, we follow the temporal sequence defined by the \texttt{LIDAR\_TOP} sensor and obtain a set of ego poses
$\{\mathcal{P}_i\}_{i=1}^N$.
Each pose $\mathcal{P}_i$ consists of a 3D translation
$\mathbf{t}_i = (x_i, y_i, z_i)^\top$, a unit-quaternion rotation
$\mathbf{q}_i$, and a timestamp $t_i$.
Within each nuScenes location (e.g., \texttt{boston-seaport}), the global coordinates provide a consistent, locally planar map in which horizontal translations $(x_i, y_i)$ describe the vehicle's trajectory.

\paragraph{Geodetic reference and modeling assumptions.}
To relate this local map to Earth-centered geographic coordinates, we introduce a reference latitude–longitude pair
$(\phi_{\mathrm{ref}}^\ell, \lambda_{\mathrm{ref}}^\ell)$
for each location $\ell$.
The reference point is chosen near the south-western boundary of the nuScenes map and serves as the anchor of a local tangent plane.
Such planar georegistration is a common convention for urban-scale coordinate systems, and because each nuScenes region spans only a few thousand meters, the induced geodesic distortion is negligible.

\paragraph{From planar displacements to geodetic coordinates.}
Given an ego pose $\mathcal{P}_i$, its horizontal translation $(x_i, y_i)$ is treated as a displacement on the local tangent plane.
We compute its ground-plane distance
\begin{equation}
    d_i = \sqrt{x_i^2 + y_i^2},
\end{equation}
and its azimuth using the two-argument arctangent
\begin{equation}
    \theta_i = \operatorname{atan2}(x_i, y_i),
\end{equation}
which correctly resolves the angular quadrant.
Starting from the reference coordinate
$(\phi_\mathrm{ref}^\ell, \lambda_\mathrm{ref}^\ell)$,
we apply a forward geodesic model
\begin{equation}
    (\phi_i, \lambda_i)
    = F\!\bigl(\phi_\mathrm{ref}^\ell, \lambda_\mathrm{ref}^\ell, \theta_i, d_i\bigr),
\end{equation}
where $F(\cdot)$ performs the standard “move by distance $d_i$ along initial bearing $\theta_i$’’ operation on a WGS-84 ellipsoid.
Within the spatial extent of nuScenes, the planar distance $d_i$ provides an accurate approximation to the geodesic arc length.

\paragraph{Per-frame localization for geographic retrieval.}
Applying this transformation independently to every ego pose yields a set of georeferenced records
$(t_i, \phi_i, \lambda_i, \mathbf{q}_i)$.
The latitude–longitude coordinates $(\phi_i, \lambda_i)$ serve as query points for the Google Maps APIs, enabling frame-level association between nuScenes and geographic data.

\subsection{Geographic Data Retrieval from Google Maps APIs}
\label{sec:supp_gmaps_retrieval}

\subsubsection{Street View Image Acquisition}
\label{sec:supp_streetview}

\paragraph{Street View Metadata API vs.\ Image API.}
Our street view acquisition pipeline is built upon two complementary Google interfaces:
the Street View \emph{Metadata API} and the Street View \emph{Image API}.
Both APIs accept a geographic query location $(\phi_i, \lambda_i)$,
but they differ fundamentally in purpose and returned information.

The Metadata API returns a structured description of the street view panorama geographically closest to the query point.
Its response includes:
(i) a success status flag,
(ii) a \emph{unique panorama identifier} that labels the corresponding street view panorama,
(iii) the panorama’s canonical capture position in latitude–longitude form, and
(iv) auxiliary attributes such as capture date.
If the query location is not covered by street view data, the response explicitly indicates that no panorama is available.

By contrast, the Image API returns a single perspective street view image rendered at a specified heading, pitch, and field of view.
Its inputs include the geographic position, the viewing configuration, and the desired image resolution.
The returned image corresponds to a projection of the underlying panorama identified by the Metadata API.

Because these two APIs provide different levels of information,
we query the Metadata API first to determine whether a panorama exists
and to obtain its unique identifier and canonical capture location.
Only after this association is established do we use the Image API to request the multi-view perspective images required for panoramic reconstruction.
This separation allows us to avoid unnecessary downloads and ensures that all perspective images are consistently tied to a single, well-defined street view panorama.

\begin{figure}[ht!]
    \centering
    \includegraphics[width=0.9\textwidth]{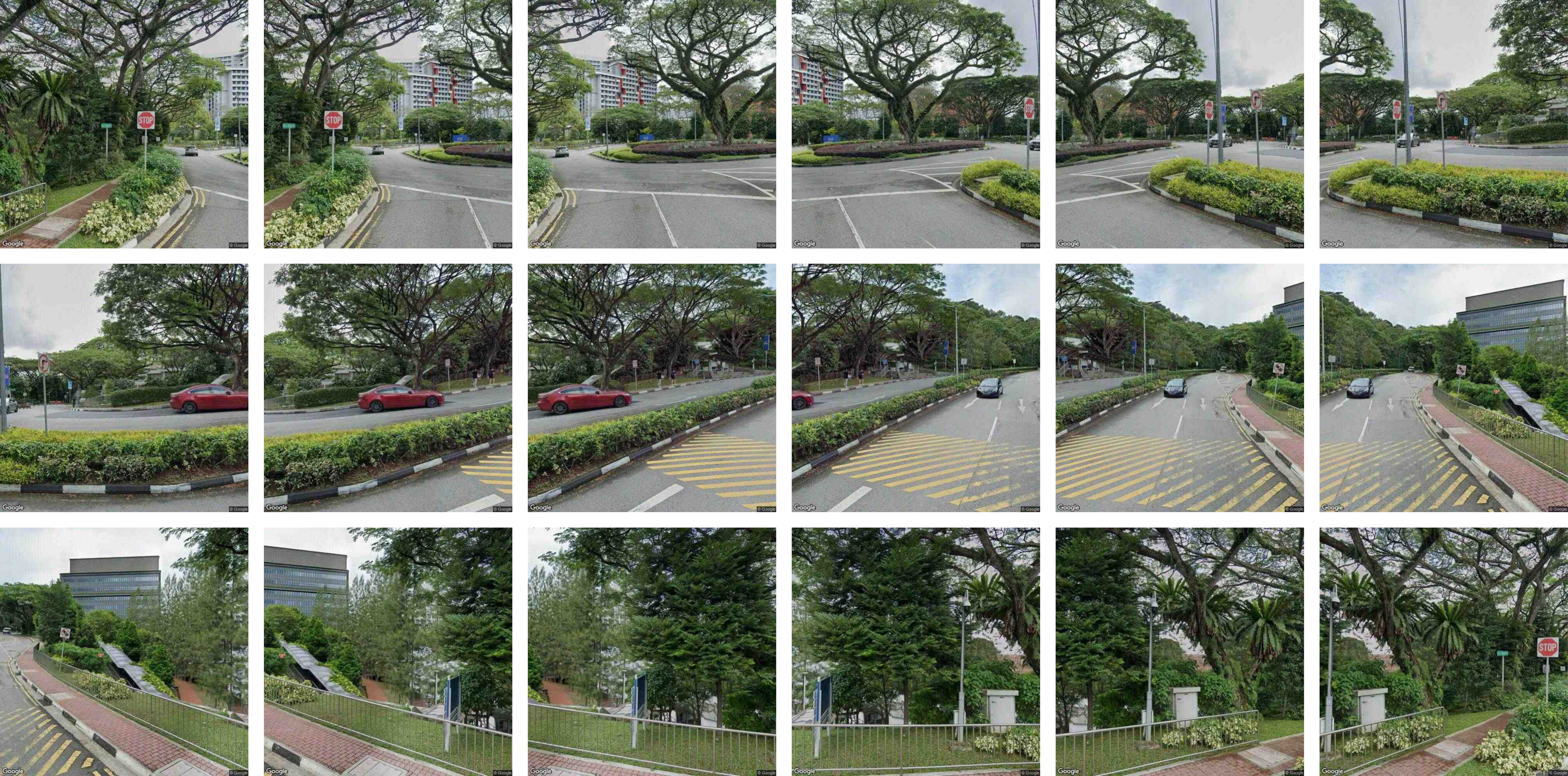}
    \vspace{-0.1mm}
    \caption{
        \textbf{Examples of Street View API outputs.} Eighteen sample images returned by the Street View APIs, illustrating panorama lookup and the multi-view perspective captures used for equirectangular reconstruction.
    }
    \vspace{-0.5mm}
    \label{fig:dataset-api-returns} 
\end{figure}

\paragraph{Metadata-driven panorama association and nearest-neighbor behavior.}
The Metadata API implicitly performs a nearest-neighbor search over Google’s street view graph:
each nuScenes coordinate $(\phi_i, \lambda_i)$ is mapped to the geographically closest existing street view panorama, if any.
This produces a deterministic association
\[
(\phi_i, \lambda_i) \;\longrightarrow\; \text{panorama index } \mathcal{P}_i.
\]
However, because the street view sampling is sparse and road networks contain parallel lanes, service roads, and split-level structures,
the nearest panorama may occasionally lie on a different road segment despite being close in Euclidean distance.
These mismatches constitute a well-defined misaligned mode and motivate the reliability analysis presented later in Sec.~\ref{sec:supp_quality}.

\paragraph{Unique panorama retrieval and caching.}
Once a panorama index $\mathcal{P}$ is identified by the Metadata API,
we ensure that its underlying panoramic content is downloaded only once.
When a panorama is encountered for the first time,
we issue a set of Image API requests at the panorama’s canonical capture location to obtain the multi-view images required for reconstructing its equirectangular representation.
The reconstructed panorama is stored on disk and indexed by its identifier.
Any subsequent frame whose Metadata query maps to the same $\mathcal{P}$ directly reuses the cached equirectangular panorama, eliminating redundant API calls.
This mechanism exploits the natural many-to-one relationship between nuScenes frames and street view panoramas while maintaining computational and storage efficiency.

\paragraph{Multi-view sampling for equirectangular reconstruction.}
To reconstruct the full $360^\circ$ panorama associated with $\mathcal{P}$,
we sample a fixed set of uniformly spaced heading angles,
\[
\Theta = \{0^\circ, 20^\circ, 40^\circ, \ldots, 340^\circ\},
\]
and request a perspective street view image for each heading at the panorama’s capture position.
As illustrated in Fig.~\ref{fig:dataset-api-returns}, every request uses a fixed horizontal field of view (60 degrees in our implementation), a fixed pitch configuration, and a fixed output resolution.
Collectively, these multi-view images provide dense angular coverage around the street view camera and serve as input for the equirectangular projection and stitching process described in Sec.~\ref{sec:supp_panorama}.

\paragraph{Frame-to-panorama mapping structure.}
The combination of the Metadata API and Image API yields a compact, surjective mapping
\[
\text{nuScenes frame } i \;\longrightarrow\; \text{panorama } \mathcal{P}_i,
\]
where many nuScenes frames may share the same panorama.
This mapping guarantees that each nuScenes frame is associated with a unique equirectangular street view panorama, and that all perspective renderings used in later stages (e.g., virtual camera synthesis) are derived from a single, geometrically consistent spherical representation.

\subsubsection{Satellite images and Map Data}
\label{sec:supp_satellite}

\paragraph{Overview.}
In addition to the street view panoramas, we incorporate high-resolution satellite images aligned with the nuScenes maps.
Satellite images provides a static top-down representation of the environment and does not require the spherical modeling or equirectangular reconstruction used for street view.

\paragraph{Region of interest and spatial resolution.}
For each nuScenes location $\ell$, we determine the spatial extent of its map in the local coordinate frame and anchor this extent to the geodetic reference point
$(\phi_{\mathrm{ref}}^\ell, \lambda_{\mathrm{ref}}^\ell)$
introduced in Sec.~\ref{sec:supp_coordinates}.
From this anchor, a geographic bounding region is defined that fully covers all trajectories within the corresponding nuScenes split.
Satellite images for this region is requested at a ground sampling distance of
\[
0.15 \text{ meters per pixel},
\]
ensuring that each pixel in the resulting raster corresponds to approximately \(0.15\,\mathrm{m}\) on the ground.

\paragraph{Single-raster satellite mosaic per location.}
All satellite tiles covering the defined geographic region are aggregated into a single georeferenced mosaic for each location $\ell$.
This mosaic fully spans the nuScenes map footprint and serves as the unified satellite representation for all frames belonging to that location.

\paragraph{Per-frame, pose-aware satellite crops.}
For a nuScenes frame with geodetic coordinates $(\phi_i, \lambda_i)$ and an ego-vehicle orientation, we compute a pose-aligned satellite crop in two steps.
First, the frame’s geodetic position is transformed into pixel coordinates on the mosaic using the known geographic reference.
This is a direct affine mapping from latitude–longitude to image coordinates.
Second, a fixed-size satellite patch is extracted around the mapped pixel and rotated according to the ego-vehicle yaw angle.
The rotation aligns the vehicle’s forward direction with the \emph{right-hand side} of the cropped satellite patch.
Thus, each crop is both spatially centered at the vehicle location and oriented in accordance with the vehicle’s heading, yielding a canonical, pose-aware top-down representation.

\subsection{Equirectangular Panorama Construction}
\label{sec:supp_panorama}

\paragraph{Motivation.}
Each street view panorama is conceptually a spherical environment map centered at the street view camera position.
However, the Google Street View Image API only provides perspective renderings of this sphere at user-specified headings and pitch angles.
To obtain a complete and consistent representation of the underlying spherical scene, we reconstruct an equirectangular panorama from multiple perspective views sampled around the panorama center.

\paragraph{Multi-view sampling of the street view sphere.}
Given the canonical capture location returned by the Metadata API, we sample the street view sphere at a fixed set of uniformly spaced heading angles,
\[
\Theta = \{0^\circ, 20^\circ, \ldots, 340^\circ\},
\]
and at a fixed pitch configuration.
Each sample corresponds to a perspective projection with a fixed horizontal field of view and image resolution.
These images form a collection of overlapping observations covering the full $360^\circ$ horizontal field around the street view camera.

\paragraph{Spherical projection model.}
To integrate these perspective views into a single panorama, each image is interpreted as a collection of rays originating from the panorama center.
For a pixel at coordinates $(u, v)$ in a perspective image with known intrinsics,
the corresponding 3D ray direction $\mathbf{d}$ is computed through inverse camera projection.
This ray is then parameterized on the unit sphere using longitude–latitude angles
\[
(\theta, \phi) \in [-\pi, \pi) \times [-\frac{\pi}{2}, \frac{\pi}{2}],
\]
where $\theta$ is the azimuth and $\phi$ is the elevation.
The equirectangular panorama is defined on a regular grid over $(\theta, \phi)$, and each incoming ray contributes its color to the corresponding grid cell.

\paragraph{Equirectangular projection and rasterization.}
The spherical surface is discretized into a raster.
For each heading–pitch configuration in the sampled set, the corresponding perspective image is projected onto this raster.
This projection uses a forward-mapping formulation:
the spherical direction $(\theta, \phi)$ is mapped to pixel coordinates
\[
x = \frac{W}{2\pi}(\theta + \pi), \qquad
y = \frac{H}{\pi}\left(\frac{\pi}{2} - \phi\right),
\]
and the color is written into the equirectangular grid.
A binary mask records which pixels receive valid contributions, ensuring that overlap regions from multiple views can be correctly combined.

\begin{figure}[h!]
    \centering
    \includegraphics[width=0.9\textwidth]{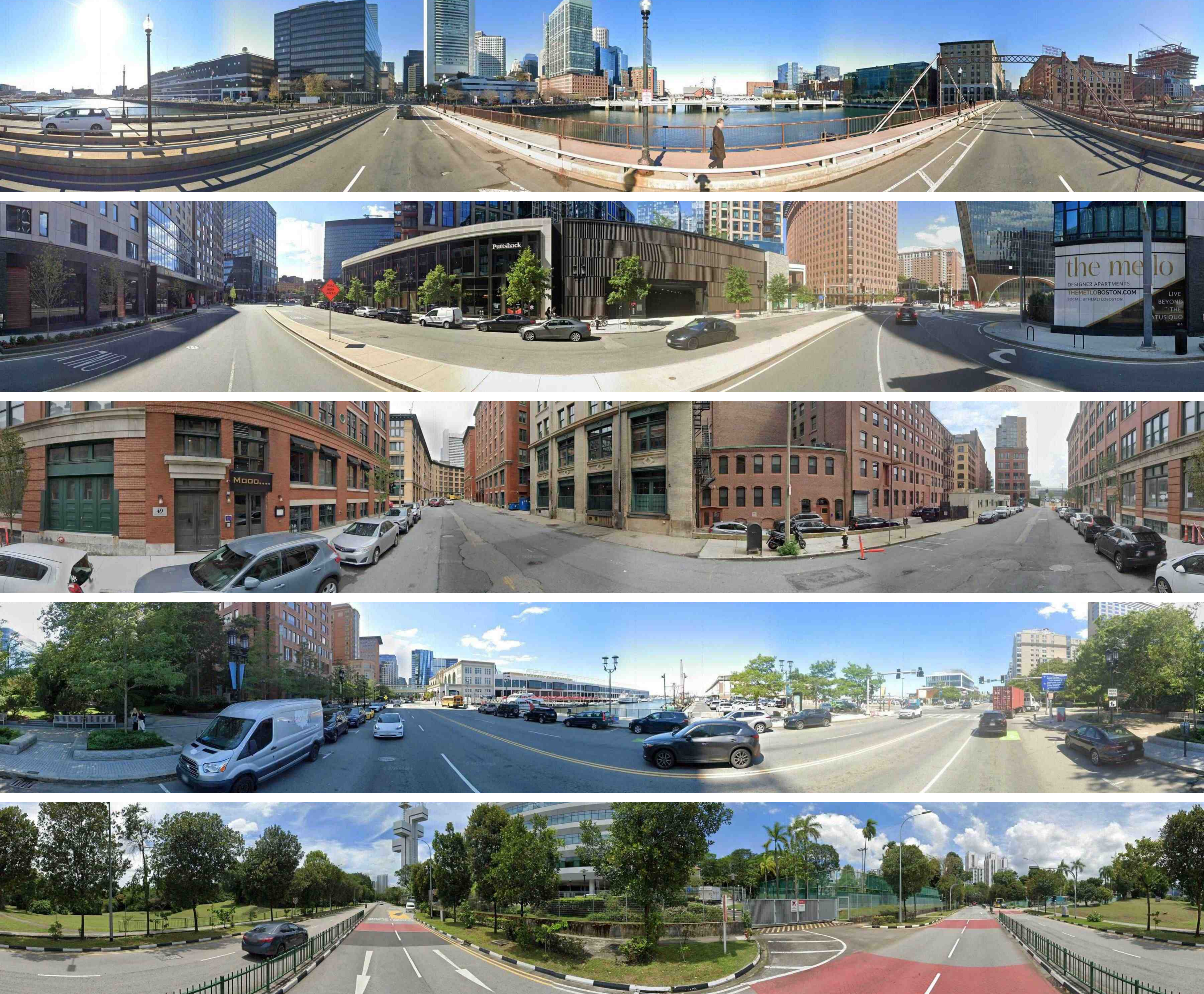}
    \vspace{-0.1mm}
    \caption{
        \textbf{Examples of reconstructed equirectangular panoramas.} Five representative panoramas generated using our multi-view sampling procedure. 
        Each panorama is produced by aggregating multiple overlapping street view perspective images sampled at uniformly spaced heading angles.
        The resulting equirectangular representations are dense and seamless, serving as the canonical environment map for subsequent virtual camera synthesis.
    }
    \vspace{-0.5mm}
    \label{fig:dataset-equirectangular} 
\end{figure}

\paragraph{Mask-based panoramic stitching.}
The equirectangular panorama is assembled by compositing all projected views.
For each view, only pixels whose rays fall within the valid forward-facing region are retained; invalid regions (e.g., back-facing directions or watermark areas) are discarded using the mask.

\paragraph{Resulting spherical representation.}

This produces a dense, seamless panorama encoding the full horizontal surround view, as illustrated in Fig.~\ref{fig:dataset-equirectangular}.

This panorama serves as the unique, spherical representation associated with each street view location and constitutes the basis for synthesizing nuScenes-aligned virtual camera views in Sec.~\ref{sec:supp_virtual_camera}.

\begin{figure}[]
    \centering
    \includegraphics[width=\linewidth]{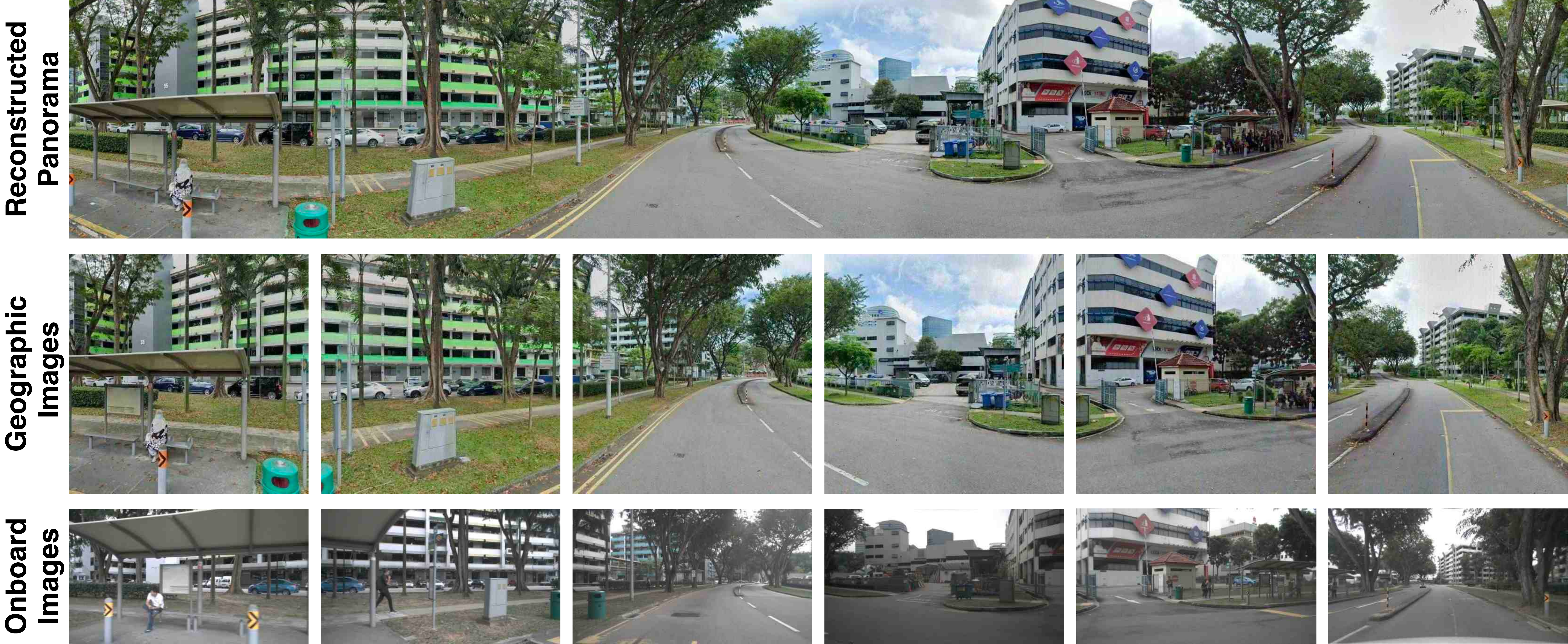}
    \vspace{-1mm}
    \caption{
        \textbf{Panorama projection and corresponding onboard views.}
        The top row shows the equirectangular panorama associated with a street view location. 
        The middle row presents six perspective views obtained by projecting the panorama 
        through the virtual camera model in Sec.~\ref{sec:supp_virtual_camera}. 
        The bottom row displays the corresponding six nuScenes onboard camera images. 
        These examples illustrate that the projection and reprojection steps provide 
        a well-aligned geometric correspondence across the panorama domain and the nuScenes camera domain.
    }
    \label{fig:supp-dataset-reproj}
    \vspace{-2mm}
\end{figure}

\subsection{Virtual Camera Model and Panorama Reprojection}
\label{sec:supp_virtual_camera}

\paragraph{Overview.}
To obtain street view images that matches the viewing geometry of nuScenes camera frames, we construct for each frame a virtual camera located at the street view panorama position.
Its orientation follows the calibrated nuScenes camera, while its intrinsic parameters follow a unified perspective model consistent with the rendering geometry used during panorama extraction.
A view aligned with this virtual camera is then synthesized by sampling the panorama.

\paragraph{Intrinsic model.}
All virtual cameras share a fixed horizontal field of view (approximately $60^\circ$) and a square output resolution $(W,H)$.
The focal length is defined as
\begin{equation}
    f_x = \frac{W}{2\tan(\mathrm{FOV}/2)}, \qquad
    f_y = f_x,
\end{equation}
with the principal point placed at the image center,
\begin{equation}
    c_x = \frac{W}{2}, \qquad c_y = \frac{H}{2}.
\end{equation}
The resulting intrinsic matrix is
\begin{equation}
    K =
    \begin{pmatrix}
        f_x & 0   & c_x \\
        0   & f_y & c_y \\
        0   & 0   & 1
    \end{pmatrix}.
\end{equation}
This formulation matches the projection geometry used for the panorama construction and ensures consistent spherical--perspective mapping.

\paragraph{Extrinsic construction.}
The virtual camera is positioned at the street view capture location associated with each nuScenes frame.
The displacement between the nuScenes ego location and the street view panorama is computed in geodetic coordinates and converted into a metric offset on a local plane.
The vertical component is set to a constant height of $2\,\mathrm{m}$.
To express this translation in the ego-vehicle coordinate system, the offset is rotated by the inverse of the ego orientation.
The rotation of the virtual camera is taken from the calibrated nuScenes camera for the corresponding sensor.
Together, these parameters define a $4\times4$ camera pose used for spherical reprojection.

\paragraph{Inverse projection to rays.}
For each output pixel $(u,v)$, the corresponding camera ray is computed using the intrinsic matrix.
Let
\begin{equation}
    X = \frac{u-c_x}{f_x}, \qquad
    Y = \frac{v-c_y}{f_y}, \qquad
    Z = 1,
\end{equation}
and normalize to obtain a unit direction:
\begin{equation}
    \mathbf{d}(u,v) =
    \frac{1}{\sqrt{X^2 + Y^2 + Z^2}}
    \begin{pmatrix}
        X \\ Y \\ Z
    \end{pmatrix}.
\end{equation}
Applying the virtual camera rotation yields the ray direction in the panorama coordinate frame.

\begin{figure}[]
    \centering
    \includegraphics[width=0.7\linewidth]{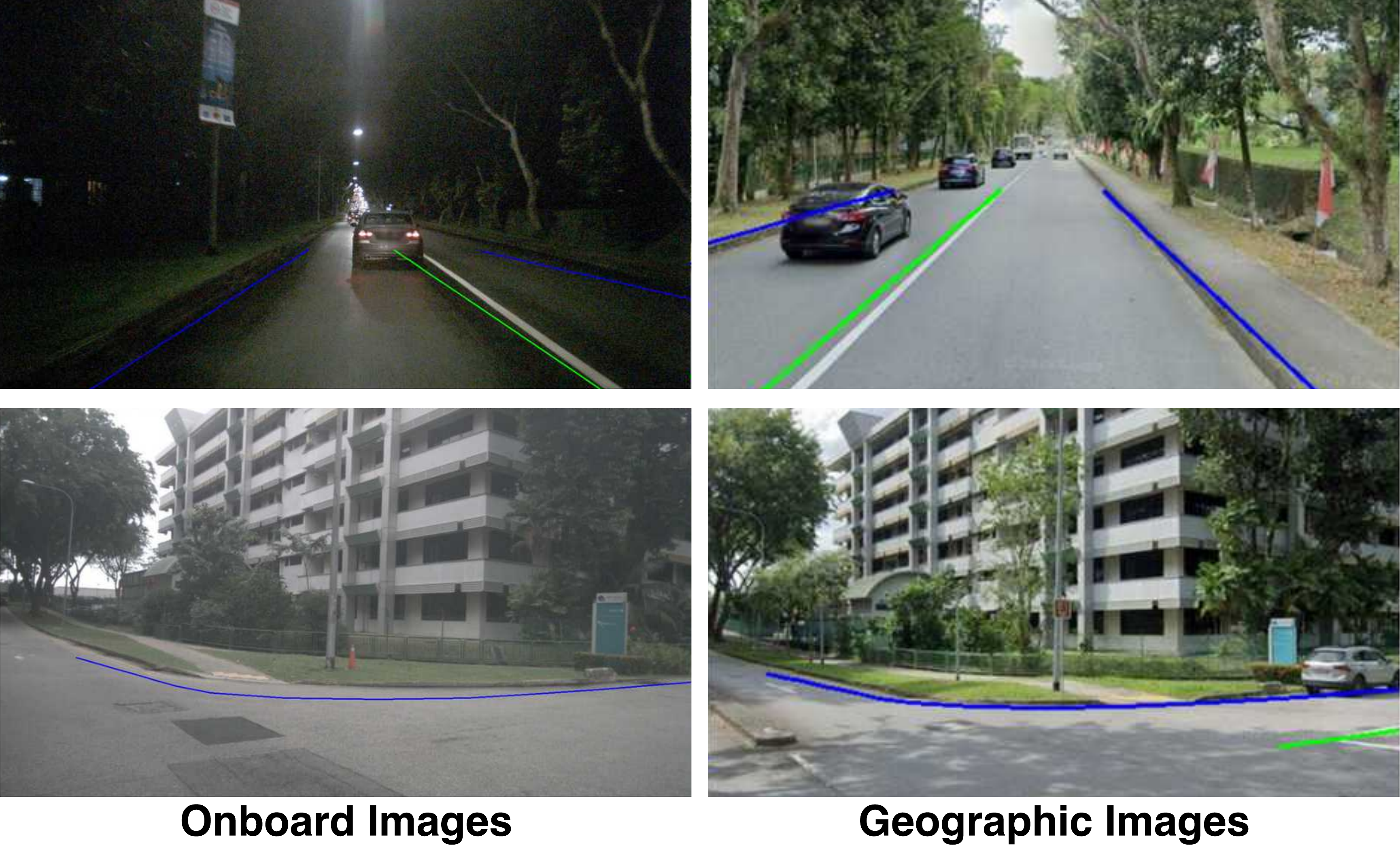}
    \vspace{-0.5mm}
    \caption{
    \textbf{Geometric consistency of lane reprojection.}
    Visualization of structural annotations reprojected into the aligned street view image using the virtual camera model in Sec.~\ref{sec:supp_virtual_camera}.
    The observed spatial alignment between the overlaid geometric curves and the roadway appearance suggests that the overall mapping and reprojection pipeline produces a well-aligned geometric correspondence across domains.
    }
    \vspace{-1mm}
    \label{fig:supp-dataset-line}
\end{figure}

\paragraph{Spherical mapping and sampling.}
Given the rotated ray $\mathbf{d}' = (X',Y',Z')^\top$, we compute its spherical azimuth and elevation as
\begin{equation}
    \theta = \arctan2(X', Z'), \qquad
    \phi = \arcsin(Y').
\end{equation}
For an equirectangular panorama of width $W_{\mathrm{pano}}$ and height $H_{\mathrm{pano}}$, the corresponding pixel coordinates are
\begin{equation}
    x = \frac{\theta + \pi}{2\pi} \, W_{\mathrm{pano}}, \qquad
    y = \frac{\tfrac{\pi}{2}-\phi}{\pi} \, H_{\mathrm{pano}}.
\end{equation}
Sampling the panorama at $(x,y)$ produces the color of pixel $(u,v)$ in the synthesized street view image.

\paragraph{Aligned street view viewpoints.}
The resulting image represents the street view environment as observed from the nuScenes camera orientation, as shown in Fig.~\ref{fig:supp-dataset-reproj}.
By applying this process to each nuScenes frame, we obtain a geometrically aligned street view counterpart for every camera observation, providing a consistent basis for spatial correspondence and cross-modal retrieval.
As shown in Fig.~\ref{fig:supp-dataset-line}, this projection mechanism produces a geometrically accurate alignment between the reprojected structures and the street view imagery.

\begin{figure}[h!]
    \centering
    \includegraphics[width=0.9\linewidth]{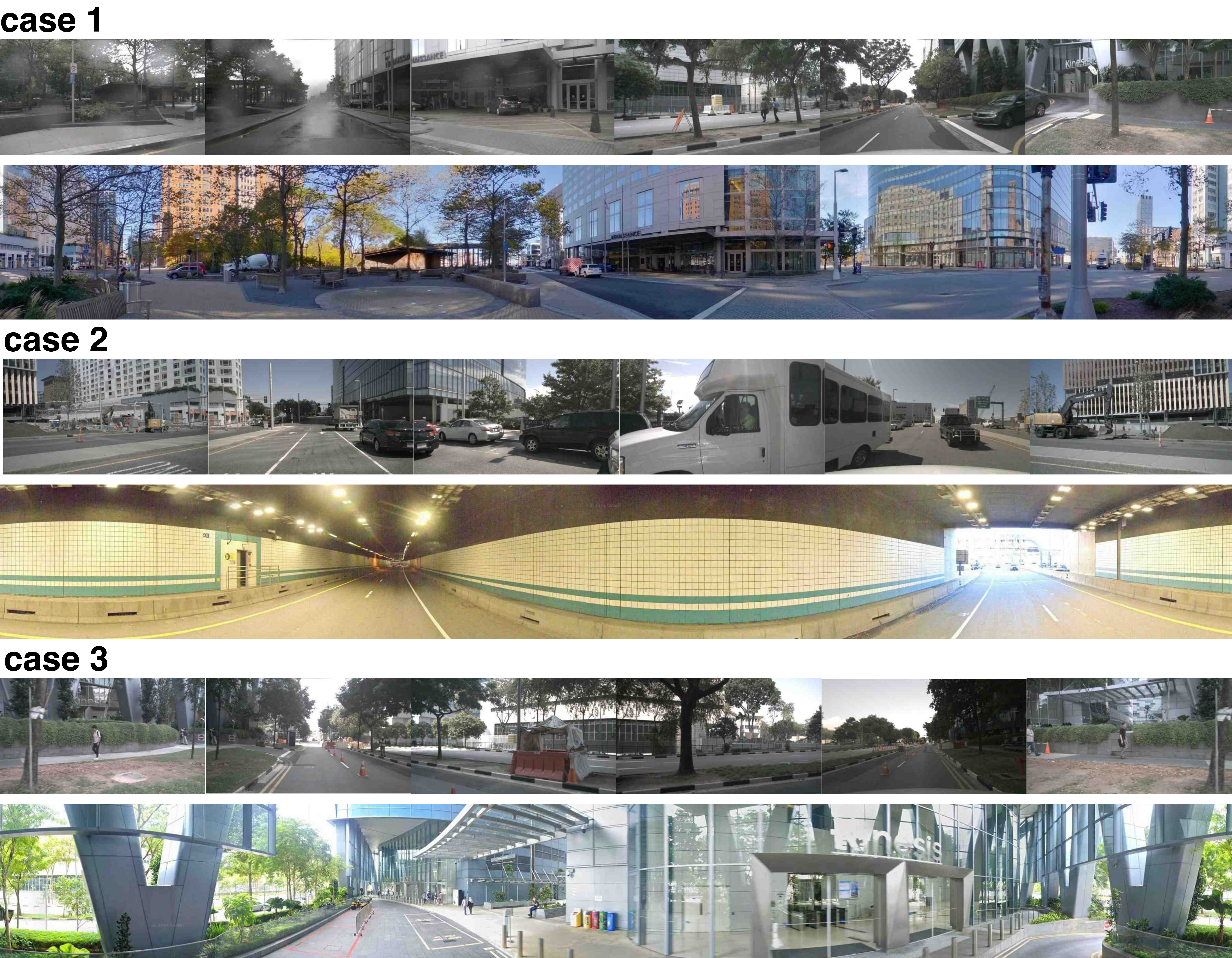}
    \vspace{-1.5mm}
    \caption{
        \textbf{Representative misaligned cases of API-returned panoramas.}
        Each row shows a mismatch example, where the \emph{top} image is the corresponding nuScenes onboard camera view and the \emph{bottom} image is the retrieved street view panorama. 
        From top to bottom, the three cases illustrate: 
        (1) a non-road panorama returned for an outdoor driving frame, 
        (2) an incorrect vertical level due to bridge--ground confusion, 
        (3) a parallel-road mismatch where the panorama lies on an adjacent roadway.
    }
    \label{fig:supp-dataset-negative}
    \vspace{-1mm}
\end{figure}

\subsection{Retrieval Misalignment and Manual Reliability Annotation}
\label{sec:supp_retrieval_annotation}

\paragraph{Overview.}
Although the Google Street View Metadata API performs a nearest-neighbor search in geographic space and typically returns the panorama closest to a queried nuScenes location, the retrieved panorama is not guaranteed to correspond to the viewpoint captured by the nuScenes camera. Due to the sparsity and heterogeneous spatial distribution of street view coverage, mismatches arise in several systematic forms. To ensure the integrity of our aligned dataset and to enable supervised learning of retrieval reliability, we manually annotate all API-returned panoramas associated with nuScenes frames.

\paragraph{Misaligned modes of API-returned panoramas.}
In practice, incorrect panorama associations arise from the following well-defined misaligned modes:
\begin{itemize}
    \item \textbf{Indoor or non-road panoramas.}
    The API may return panoramas captured in buildings, parking structures, or enclosed environments, even when the nuScenes frame is recorded on a public road.
    
    \item \textbf{Incorrect vertical level (bridge--ground confusion).}
    For multi-level infrastructure, the nearest panorama may lie at the wrong elevation, showing the underside of a bridge when the vehicle is driving above it, or the reverse.
    
    \item \textbf{Parallel-road ambiguity.}
    Dense urban areas often contain parallel roads only a few meters apart. Proximity-based retrieval may incorrectly return a panorama from the adjacent roadway, leading to consistent semantic and structural mismatch.
    
\end{itemize}

As shown in Fig.~\ref{fig:supp-dataset-negative}, these misalignment patterns constitute the primary sources of retrieval noise in the nuScenes–street-view pairing, motivating the need for a systematic annotation procedure.

\paragraph{Manual reliability annotation.}
Each retrieved panorama is manually compared with the corresponding nuScenes frame. Annotators inspect the panorama in its reconstructed equirectangular form and the nuScenes image in its native camera geometry. The annotation task is strictly comparative and relies only on observable scene-level cues, including roadway configuration, building geometry, structural alignment, and global viewing direction.

\vspace{1mm}

\begin{figure}[]
    \centering
    \includegraphics[width=0.95\linewidth]{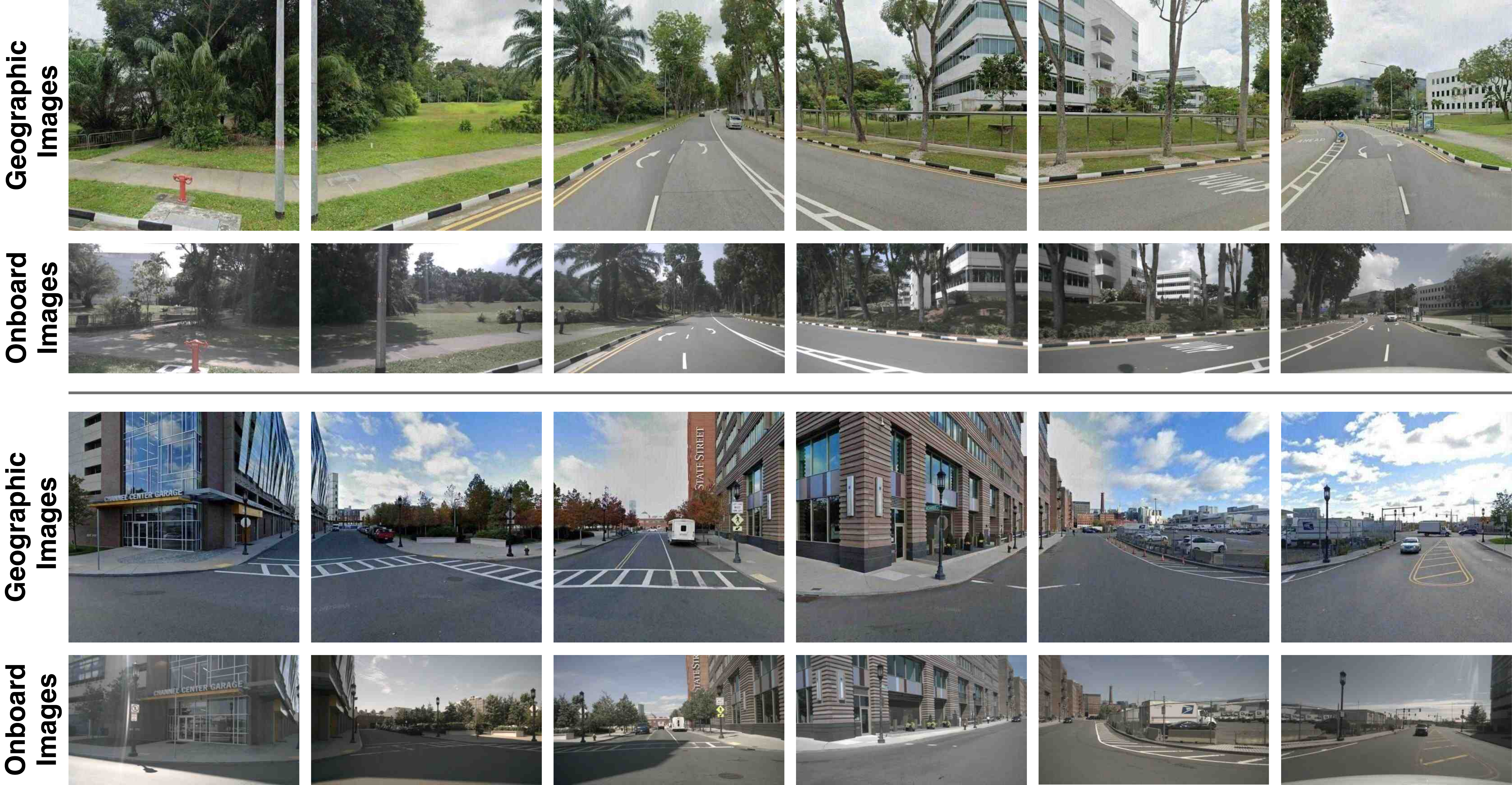}
    \vspace{-1.5mm}
    \caption{
        \textbf{Examples of reliable positive matches.}
        The figure presents two positive cases, each consisting of a street view panorama (top) and its corresponding nuScenes onboard camera image (bottom). 
        In both cases, the two modalities exhibit consistent scene structure, including roadway layout, surrounding buildings, and overall orientation. 
        These clear and coherent correspondences are used as reliable positive supervision in our analysis.
    }
    \label{fig:supp-dataset-positive}
    \vspace{-1mm}
\end{figure}

To provide clean and unambiguous supervision, we adopt a ternary labeling scheme:
\begin{itemize}
    \item \textbf{Reliable (positive).}
    Assigned when the panorama and the nuScenes image exhibit clear semantic and geometric consistency, with matching road layout, scene structure, and approximate orientation, as shown in Fig.~\ref{fig:supp-dataset-positive}.

    \item \textbf{Unreliable (negative).}
    Assigned when the panorama clearly corresponds to a different physical environment or viewpoint, such as incorrect elevation, adjacent roads, or indoor locations, as shown in Fig.~\ref{fig:supp-dataset-negative}.

    \item \textbf{Unlabeled (ambiguous).}
    Used for cases where the correspondence cannot be judged confidently—for example, when the panorama is geographically close but exhibits subtle structural differences not resolvable from the nuScenes view. These samples are excluded from all supervised training.
\end{itemize}

This design provides a controlled supervisory signal by retaining only cases with unambiguous alignment or misalignment, thereby ensuring learnability.

\section{Implementation Details of Neural Networks}
\label{sec:supp_model}

\subsection{3D Postional Encodings for Retrieved Geographic Images}
\label{sec:petr}

To incorporate retrieved street-view and satellite images into BEV-based
representations in a geometrically consistent manner, we adopt the positional
encoding mechanism introduced in PETR~\cite{petr}. Geographic images
provide valuable spatial priors, but their feature maps reside in the image
plane and do not contain explicit 3D information. PETR enables the construction
of geometry-aware positional encodings that describe the spatial coordinates
associated with each pixel location, thereby facilitating cross-view
interaction with BEV features.

\paragraph{Depth Sampling and Coordinate Generation.}
For each geographic image, a discrete set of depth values
\(\{d_k\}_{k=1}^{D}\) is defined within the valid operating range of the
dataset. Each pixel coordinate \((u,v)\) is then associated with a
three-dimensional point computed via inverse projection:
\[
\mathbf{x}_{uvk} = \pi^{-1}(u,v,d_k),
\]
where \(\pi^{-1}\) applies the camera intrinsics and the corresponding
street-view or satellite camera pose. This procedure generates a dense set of
3D coordinates \(\mathbf{x}_{uvk}\) for all pixels and depth samples.

\paragraph{Normalization and Positional Encoding.}
Following PETR, the 3D coordinates are normalized to a fixed range
\([x_{\min},y_{\min},z_{\min}] - [x_{\max},y_{\max},z_{\max}]\) to ensure
numerical stability:
\[
\tilde{\mathbf{x}}_{uvk} =
\frac{\mathbf{x}_{uvk}-\mathbf{x}_{\min}}
{\mathbf{x}_{\max}-\mathbf{x}_{\min}}.
\]
The depth dimension is concatenated along the channel axis, after which a
lightweight \(1\times1\) convolutional mapping is applied:
\[
\mathbf{P}_{\mathrm{geo}} = f_{\theta}(\tilde{\mathbf{x}}),
\]
producing a geometry-aware positional embedding aligned spatially with the
retrieved geographic feature map.

\paragraph{Street View and Satellite Encoding.}
For retrieved street-view images, the full PETR-style depth frustum is used,
based on the camera intrinsics and extrinsics stored in the metadata.  For satellite images, which typically corresponds to a near-planar observation of the environment, pixel locations are mapped to the LiDAR coordinate system via a specific 2D transformation, with the vertical coordinate set to $z=0$.  Both cases employ the same positional encoder
\(f_{\theta}\), allowing a unified geometric representation across modalities.

\paragraph{Integration into Cross-Attention.}
The positional encoding \(\mathbf{P}_{\mathrm{geo}}\) serves as the
geometry-aware key in the Geo-Cross-Attention module:
\[
\mathrm{CrossAttn}(
\mathbf{F}_{\mathrm{BEV}},
\mathbf{F}_{\mathrm{geo}} + \mathbf{P}_{\mathrm{geo}},
\mathbf{F}_{\mathrm{geo}}
),
\]
enabling BEV queries to attend to retrieved geographic features according to
their spatial correspondence rather than solely on appearance. This
positional encoding ensures that retrieved geographic features are integrated
into BEV space in a manner consistent with the underlying 3D structure.

\subsection{Spatial Retrieval Adapter}

Our Spatial Retrieval Adapter is designed to be highly flexible and can be integrated into various BEV perception models. 

\paragraph{BEVFormer-based Architecture}
For architectures that are inherently attention-based, such as BEVFormer~\cite{bevformer}, the integration is particularly natural. The BEVFormer encoder consists of stacked layers, each performing temporal self-attention and spatial cross-attention. We simply insert our \textbf{Geo-Cross-Attention} module at the end of each of these layers. This allows the model to progressively refine its BEV representation at each stage of the encoder, first using onboard sensor data and then enriching it with the retrieved global context, while keeping the rest of the architecture unchanged.

\paragraph{Other BEV Architectures.}
The adapter's plug-and-play nature allows for flexible application in other models, such as those based on Lift-Splat-Shoot (LSS)~\cite{lss}. The core principle is to apply the Geo-Cross-Attention module at any stage where a BEV feature map is present. We demonstrate this flexibility with two examples:
\begin{itemize}
    \item For \textbf{BEVDet}~\cite{bevdet}, which processes BEV features at multiple scales, we apply our adapter after each downsampling stage in the BEV backbone. This enables multi-scale fusion of the geographic context.
    \item For \textbf{Flash-Occ}~\cite{yu2023flashocc}, to accommodate its significant GPU memory requirements, we apply the Geo-Cross-Attention module only once, on the final and most compact BEV feature map (100$\times$100 resolution). This highlights the adaptability of our method to different computational and memory constraints.
\end{itemize}

\paragraph{Generative Architecture.} Existing generative world models~\cite{bench2driveR,gao2024magicdrivedit} for autonomous driving are mostly built by fine-tuning pretrained generative models~\cite{sd3}. Concretely, the original MM-DiT backbone is frozen, and trainable temporal as well as spatial cross-attention modules are added to the DiT blocks to enhance multi-view and multi-frame consistency. Our geo-adapter is inserted after the spatiotemporal cross-attention and uses zero-initialized convolutions together with gated linear layers to ensure minimal disturbance to the original model.

These examples underscore that our adapter is not tied to a specific architecture but can be strategically placed to balance performance and efficiency.

\subsection{Reliability Estimation Gate}
\label{sec:supp_quality}

Geographic retrieval unavoidably suffers from missing, outdated, or
misaligned street-view images. Such imperfections introduce uncertainty in
the external priors and can negatively affect downstream perception.
To ensure that the model incorporates geographic information only when it is
trustworthy, we employ a \emph{Reliability Estimation Gate}, which produces a
confidence score $w\in[0,1]$ that modulates the contribution of retrieved
features during fusion. The gate combines both appearance-level and
geometry-level cues, enabling the system to handle noisy geographic coverage.

\paragraph{Feature Similarity via ZNCC.}
The primary indicator of reliability is the visual consistency between current
onboard observations and the retrieved street-view features.
Given an onboard feature map $\mathbf{F}_{\mathrm{on}}$ and a retrieved feature
$\mathbf{F}_{\mathrm{geo}}$, the latter is first aligned to the onboard spatial
resolution through scale-preserving interpolation followed by center cropping.
A Zero-Normalized Cross-Correlation (ZNCC) operator is then applied locally to
measure their photometric agreement:
\[
\mathrm{ZNCC}(x,y)
=\frac{\mathbb{E}\!\left[(x-\mu_x)(y-\mu_y)\right]}
{\sqrt{\mathbb{E}[(x-\mu_x)^2]\,\mathbb{E}[(y-\mu_y)^2]}+\epsilon},
\]
where $x$ and $y$ denote corresponding patches with 9$\times$9 kernel size and $\mu_x,\mu_y$ their local
means. We convert ZNCC ranges in $[-1,1]$ into a normalized
difference measure,
\[
\mathrm{Diff}=\frac{1-\mathrm{ZNCC}}{2}\in[0,1],
\]
which provides a spatially dense signal reflecting how much the retrieved
appearance deviates from the real-time observation. Larger values indicate
less trustworthy geographic information.

\paragraph{Distance-Based Consistency.}
Street view images far from the ego location are more likely to reflect a
different environment or capture outdated structures.
To account for this, we encode the GPS distance $d_{\mathrm{GPS}}$ between the
retrieved viewpoint and the current position. Instead of using the raw
distance—which can become numerically unstable—we apply a bounded mapping:
\[
d' = \tanh\!\left(\frac{d_{\mathrm{GPS}}}{s}\right),
\]
where $s$ is a fixed scaling factor. This produces a distance cue $d'\in[-1,1]$
that effectively suppresses geographically irrelevant retrievals without
over-penalizing small, benign spatial offsets.

\paragraph{Reliability Prediction.}
The visual difference map and distance cue are aggregated to compute the final
reliability score:
\[
w = \sigma\left( f_{\theta}(\mathrm{Diff}, d') \right),
\]
where $f_{\theta}$ denotes a trainable prediction function and $\sigma$ is
the sigmoid activation. The resulting $w$ serves as an adaptive weight during
fusion, increasing the impact of geographically consistent retrievals and
down-weighting those that are mismatched or outdated.

\paragraph{Training Supervision.}
During training, the gate is supervised with binary labels (detailed in~\ref{sec:supp_retrieval_annotation}) indicating whether a
retrieved sample is valid (1) or invalid (0). The loss is implemented using a
binary cross-entropy objective:
\[
\mathcal{L}_{\mathrm{quality}}
= \mathrm{BCE}(w, w^\ast),
\]
while samples marked as uncertain are ignored. Through this supervision, the
gate learns to detect mismatched content, outdated appearance, or enlarged
spatial discrepancy, and to output lower reliability in these cases.

\paragraph{Behavior in Imperfect Geographic Conditions.}
The Reliability Estimation Gate ensures that the overall framework does not
become overly dependent on geographic priors. When retrievals are missing,
incomplete, or incorrect, the gate attenuates their contribution, allowing the
model to fall back smoothly on onboard sensing. The degradation studies in the
main paper demonstrate that even as the proportion of missing or perturbed
street-view inputs increases, performance degrades gracefully without
catastrophic collapse. This stability highlights the gate's ability to maintain
balanced fusion and robust perception across real-world variations in street
view density, recency, or alignment.

\section{Task Definition and Implementation Details}
\label{sec:supp_exp}

\subsection{3D Object Detection}

\paragraph{Task Definition.}

3D object detection in autonomous driving aims to localize and classify surrounding traffic participants in an ego-centric 3D coordinate system using multi-view camera inputs. Given a set of synchronized images, the model estimate the position, dimensions, orientation, and semantic category of each object in the scene. The output is typically represented as a collection of oriented 3D bounding boxes, each parameterized by a 3D center, box extents, heading angle, and class label. On nuScenes, performance is commonly quantified using mean Average Precision (mAP) and the nuScenes Detection Score (NDS), which jointly evaluate detection quality across localization, orientation, and related attributes.

We consider BEVDet~\cite{bevdet} and BEVFormer~\cite{bevformer} as baselines, which are representative multi-view BEV-based detectors on nuScenes~\cite{caesar2020nuscenes}. In our setting, geographic information is injected into their BEV feature representations via the Spatial Retrieval Adapter, while the detection heads, training objectives, and evaluation protocol follow the original works. Hyperparameters are detailed in Table~\ref{tab:hyperparams}.

\paragraph{Evaluation Metrics.}

Metrics are adopted from the official nuScenes detection challenge~\cite{caesar2020nuscenes}.
\begin{itemize}
    \item \textbf{mAP ($\uparrow$):} Mean Average Precision, calculated based on 2D center distance on the BEV plane.
    \item \textbf{NDS ($\uparrow$):} The nuScenes Detection Score, a weighted sum of mAP and five mean True Positive (mTP) metrics: average translation (ATE), scale (ASE), orientation (AOE), attribute (AAE), and velocity (AVE) errors.
    \begin{equation*}
        \text{NDS} = \frac{1}{10} \left[ 5 \cdot \text{mAP} + \sum_{\text{mTP} \in \text{TPs}} (1 - \min(1, \text{mTP})) \right]
    \end{equation*}
\end{itemize}

\subsection{Online Mapping}

\paragraph{Task Definition.}

Online mapping focuses on reconstructing local high-definition map elements that describe the static road topology needed for navigation and downstream decision making. The task takes multi-view camera images as input and predicts structured representations of static road components in an ego-centric bird’s-eye-view coordinate frame. Typical targets include lane dividers, lane boundaries, and crosswalks, which are represented as vectorized geometric primitives. The resulting map elements support lane-level reasoning and provide a compact description of the underlying road layout.

The output of this task consists of sets of polylines or curves with associated semantic categories. Because these structures are thin, elongated, and topologically constrained, online mapping is particularly sensitive to occlusion, adverse illumination, and limited sensor range. The lack of explicit depth measurements further increases the difficulty of accurately recovering road geometry from images.

We adopt MapTR~\cite{maptr} and MapTRv2~\cite{liao2025maptrv2} as baselines for this task. We conduct experiments on the nuScenes~\cite{caesar2020nuscenes} dataset. Both methods operate on BEV representations derived from multi-view images and are widely used for vectorized map prediction on nuScenes. In our experiments, geographic images aligned with the ego trajectory are encoded and fused into the BEV features through the Spatial Retrieval Adapter. The prediction heads and training pipelines of the baselines are kept unchanged. Hyperparameters are detailed in Table~\ref{tab:hyperparams}.

\paragraph{Evaluation Metrics.}
Metrics are based on the MapTR protocol for vectorized map element construction on the nuScenes dataset.
\begin{itemize}
    \item \textbf{AP ($\uparrow$):} Average Precision is used to evaluate map construction quality. Unlike standard object detection which uses IoU, MapTR utilizes Chamfer distance ($D_{Chamfer}$) to determine whether a predicted map element matches the ground truth. The final AP is calculated by averaging the APs obtained under several matching thresholds $\tau \in T$, where $T = \{0.5m, 1.0m, 1.5m\}$.
    \begin{equation*}
        \text{AP} = \frac{1}{|T|} \sum_{\tau \in T} \text{AP}_\tau
    \end{equation*}
    \item \textbf{mAP ($\uparrow$):} The mean Average Precision is the average of the calculated APs over the three map element classes: pedestrian crossing, lane divider, and road boundary ($N_{cls}=3$).
    \begin{equation*}
        \text{mAP} = \frac{1}{N_{cls}} \sum_{i=1}^{N_{cls}} \text{AP}_i
    \end{equation*}
\end{itemize}

\subsection{Occupancy Prediction}

\paragraph{Task Definition.}

3D occupancy prediction formulates scene understanding as a volumetric classification problem, where the environment around the ego vehicle is discretized into a 3D grid and each voxel is assigned an occupancy state and a semantic label. The task aims to capture both static infrastructure and dynamic participants in a dense and geometrically explicit representation. This volumetric view of the scene is particularly useful for downstream planning and simulation, as it provides fine-grained information about which regions of space are free, traversable, or blocked.

The model takes multi-view camera images as input, aggregates them into a BEV feature representation, and decodes a multi-class 3D occupancy grid over a fixed spatial range. The Occ3D-nuScenes benchmark~\cite{tian2023occ3d} reports mean Intersection-over-Union (mIoU) over semantic classes, including driveable area, other flat, sidewalk, terrain, manmade structures, vegetation, and dynamic categories. The task is challenging because large portions of the 3D volume may be unobserved or only partially visible, requiring the model to infer plausible completions based on structural priors and multi-view consistency.

We use FB-OCC~\cite{li2023fb} and FlashOCC~\cite{yu2023flashocc} as baselines, which are representative BEV-based occupancy prediction models evaluated on Occ3D-nuScenes~\cite{tian2023occ3d} in our experiments. Both methods lift multi-view features into a BEV space and decode them into voxel-wise predictions. In our spatial retrieval setting, geographic images provide additional information about static background layout and are fused with the BEV features through the Spatial Retrieval Adapter. The occupancy heads and loss functions remain identical to the original implementations. Hyperparameters are detailed in Table~\ref{tab:hyperparams}.

\paragraph{Evaluation Metrics.}
Metrics follow the Occ3D-nuScenes benchmark for voxel-level semantic prediction.
\begin{itemize}
    \item \textbf{mIoU ($\uparrow$):} The mean Intersection-over-Union (IoU) over $N_{cls}$ classes, where TP, FP, and FN are the counts of true positive, false positive, and false negative voxels for a given class.
    \begin{equation*}
        \text{IoU} = \frac{\text{TP}}{\text{TP} + \text{FP} + \text{FN}}, \quad \text{mIoU} = \frac{1}{N_{cls}} \sum_{i=1}^{N_{cls}} \text{IoU}_i
    \end{equation*}
\end{itemize}

\subsection{End-to-End Planning}

\paragraph{Task Definition.}

End-to-end autonomous driving planning aims to generate a safe and kinematically feasible future trajectory for the ego-vehicle directly from raw sensor inputs, conditioned on high-level navigation commands (e.g., turn left, turn right, or go straight). Unlike modular pipelines that rely on intermediate perception outputs like bounding boxes or HD maps, end-to-end approaches optimize the trajectory generation jointly. The model takes multi-view camera images as input to encode the scene context and outputs a sequence of future waypoints $\mathcal{T}_{pred} = \{p_t\}_{t=1}^{T}$ representing the ego-vehicle's planned path over a future horizon $T$.

In our experiments, we evaluate the planning performance on the nuScenes~\cite{caesar2020nuscenes} dataset using VAD~\cite{vad} as the baseline, which is a representative vectorized planning framework. VAD explicitly models the driving scene using vectorized agent motions and map elements to impose planning constraints, ensuring safety and compliance. In our spatial retrieval setting, we augment the BEV encoder of VAD by adding additional cross attention between the BEV feature and the offline retrieved geographic features through the Spatial Retrieval Adapter. These geographic priors serve as a stable environmental reference to refine the planning trajectory. Hyperparameters are detailed in Table~\ref{tab:hyperparams}.

\paragraph{Evaluation Metrics.}
Following the specific implementation of ST-P3~\cite{STP3} and VAD~\cite{vad}, we evaluate the open-loop planning performance using the Average L2 Displacement Error and the Average Collision Rate over different time horizons ($H \in \{1s, 2s, 3s\}$).

\begin{itemize}
    \item \textbf{L2 Displacement Error (L2 $\downarrow$):} This metric reports the Average Displacement Error (ADE) accumulated over the time horizon $H$. It is calculated as the mean Euclidean distance between the predicted waypoints and ground truth waypoints:
    \begin{equation*}
        \text{L2}_H = \frac{1}{N_H} \sum_{t=1}^{N_H} \| \hat{p}_{t} - p_{t}^{GT} \|_2
    \end{equation*}
    where $N_H$ is the number of time steps within the horizon $H$.

    \item \textbf{Collision Rate (Collision \% $\downarrow$):} This metric evaluates the safety of the planned trajectory by calculating the frequency of collisions across time steps. Importantly, to account for data noise, a collision at time step $t$ is only penalized if the expert (ground truth) trajectory is collision-free at that moment. The metric is defined as the average percentage of time steps involving a valid collision:
    \begin{equation*}
        \text{Collision}_H = \frac{1}{N_H} \sum_{t=1}^{N_H} \mathbb{I}\left( \mathcal{C}(\hat{p}_t, \mathcal{O}_t) \land \neg \mathcal{C}(p^{GT}_t, \mathcal{O}_t) \right) \times 100\%
    \end{equation*}
    where $\mathcal{C}(p, \mathcal{O})$ indicates whether the ego-vehicle polygon at pose $p$ intersects with the occupancy grid $\mathcal{O}$ of other agents, and $\mathbb{I}$ is the indicator function.
\end{itemize}

\subsection{Generative World Model}

\paragraph{Task Definition.}

Generative world model for autonomous driving aims to synthesize realistic, controllable multi-view driving videos conditioned on structured scene information. Given a short history of surround-view camera frames together with ego-vehicle actions, agent configurations, and environment descriptors, the model learns a conditional distribution over future video sequences. The output is a set of temporally coherent RGB streams for all cameras, which should remain consistent with the specified ego motion, dynamic-agent behaviors, and high-level scene attributes. Following prior work on generative world models~\cite{gao2023magicdrive,gao2024magicdrivedit,bench2driveR}, performance is typically assessed using distribution-level video metrics such as Fréchet Inception Distance for visual fidelity, complemented by task-specific controllability measures that compare generated content against the conditioning signals (e.g., action or 3D box consistency).

We adopt Unimlvg~\cite{chen2024unimlvg} and MagicDriveDit~\cite{gao2024magicdrivedit} as a representative multi-view latent diffusion world model baseline. These models are built on video foundation models, where Unimlvg trained to generate 19-frame videos and Magicdrivedit generate 17-frame videos. During testing, Unimlvg uses the first 3 frames as reference frames, and after the first round of generation, the last 3 frames of the generated 19-frame clip are taken as reference for the next round. In total, it performs two rounds of autoregressive generation to produce 35-frame videos for evaluation. Magicdrivedit instead uses only the first frame as the reference frame and directly generates 17-frame videos.

Based on the nuScenes dataset, Unimlvg generates 150 35-frame video clips for testing, whereas Magicdrivedit, under our setup, is evaluated on approximately 2,400 video clips with a frame interval of 15.
For each generated clip, we allocate a geographic image to assist generation according to the position of its first and last frames. The detailed training parameters are listed in Table~\ref{tab:hyperparams}.

\paragraph{Evaluation Metrics.}
Metrics assess the distributional similarity between generated and real videos.
\begin{itemize}
    \item \textbf{FVD ($\downarrow$) / FID ($\downarrow$):} Fr\'echet Video/Inception Distance between the distributions of real-world data (r) and generated data (g), which are modeled as multivariate Gaussians with mean $\mu$ and covariance $\Sigma$.
    \begin{equation*}
    \text{FVD/FID} = \left\| \mu_r - \mu_g \right\|_2^2 + \text{Tr}\left(\Sigma_r + \Sigma_g - 2(\Sigma_r \Sigma_g)^{1/2}\right)
    \end{equation*}
\end{itemize}



\begin{table*}[h!]
\centering
\caption{\textbf{Training Hyperparameters for All Baselines.} We report the total batch size across all GPUs. Configurations are aligned with the original papers for a fair evaluation. UVG and MDD are generative models, and their settings reflect the video generation task. * indicates models fine-tuned on the corresponding baseline.}
\label{tab:hyperparams}
\resizebox{\textwidth}{!}{%
\begin{tabular}{@{}lccccccc|cc@{}}
\toprule
\textbf{Hyperparameter} & \textbf{MapTR} & \textbf{MapTRv2} & \textbf{FB-OCC} & \textbf{Flash-OCC} & \textbf{BEVDet} & \textbf{BEVFormer} & \textbf{VAD} & \textbf{UVG} & \textbf{MDD} \\ \midrule
Optimizer               & AdamW& AdamW& AdamW& AdamW& AdamW& AdamW& AdamW& AdamW & AdamW\\
Learning Rate           & 6e-4& 6e-4& 2e-4& 1e-4& 2e-6& 2e-6& 2e-4& 8e-5 & 8e-5\\
LR Scheduler            & CosineAnnealing& CosineAnnealing& LinearWarmup& LinearWarmup& StepDecay & CosineAnnealing& CosineAnnealing& None & LinearWarmup\\
Weight Decay            & 0.01& 0.01& 0.01& 0.01& 0.01& 0.01& 0.01& 0.01 & 0.01\\
Batch Size (Total)      & 32& 32& 32& 32& 16& 2& 8& 8& 32\\
Training Epochs         & 110& 110& 20& 24& 4* & 4* & 60& 4* & 4* \\ \midrule
GPUs Used               & \multicolumn{7}{c|}{8 $\times$ NVIDIA RTX 4090 (48GB)} & \multicolumn{2}{c}{8 $\times$ NVIDIA H800 (80GB)} \\ 
\bottomrule
\end{tabular}%
}
\end{table*}

\section{Extra Qualitative Results}
\label{sec:supp_qual}

\begin{figure}[]
    \centering
    \includegraphics[width=0.5\linewidth]{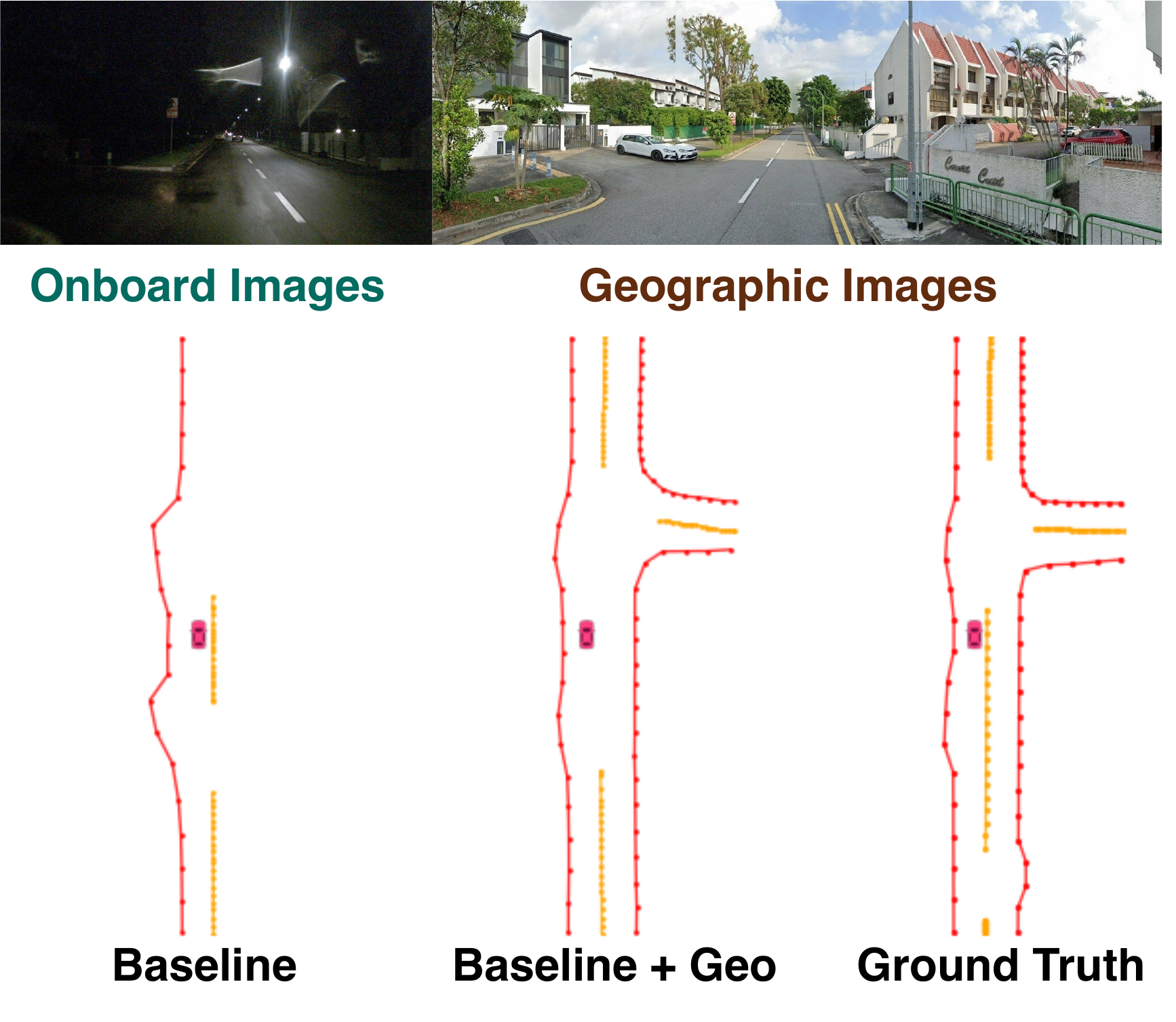}
    \caption{
        \textbf{Online Mapping:} The baseline model(MapTRv2) fails to detect the right-side lane line and intersection due to the dim lighting. Our Spatial Augmented model recovers the crucial road topology by leveraging the stable, bright static environment prior provided by the geographic images, closely matching the Ground Truth.
    }
    \label{fig:map_case1}
\end{figure}

\begin{figure}[]
    \centering
    \includegraphics[width=0.5\linewidth]{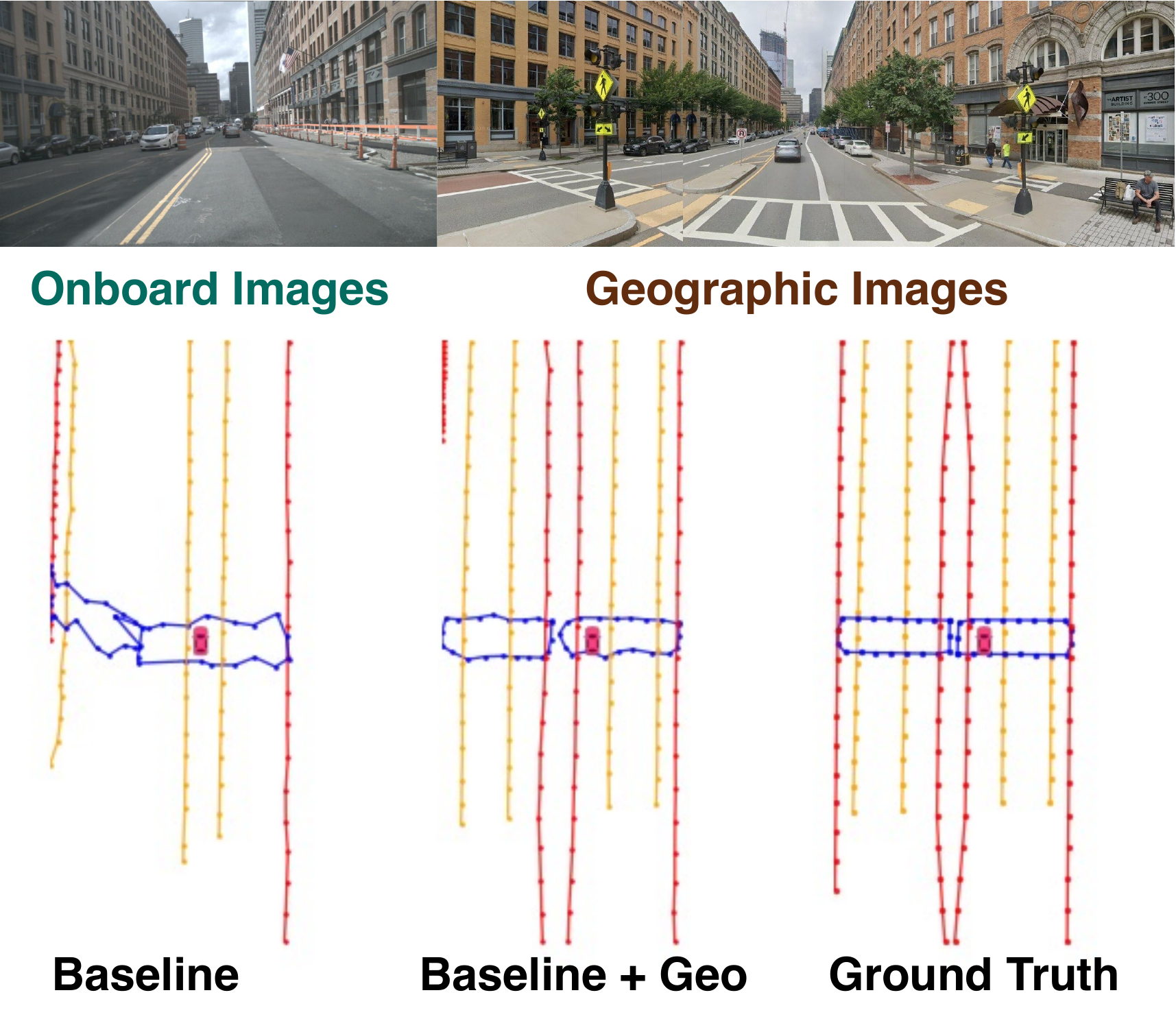}
    \caption{
        \textbf{Online Mapping:} The baseline model struggles to accurately map the pedestrian crossing (blue polygon) near the ego vehicle due to the limited field of view of onboard cameras. The non-ego-centric perspective from the geographic images provides a stable, complete spatial context, enabling our Spatial Augmented model to recover the correct geometric shape, resulting in a prediction highly consistent with the Ground Truth.
    }
    \label{fig:map_case2}
\end{figure}

\begin{figure}[]
    \centering
    \includegraphics[width=1\linewidth]{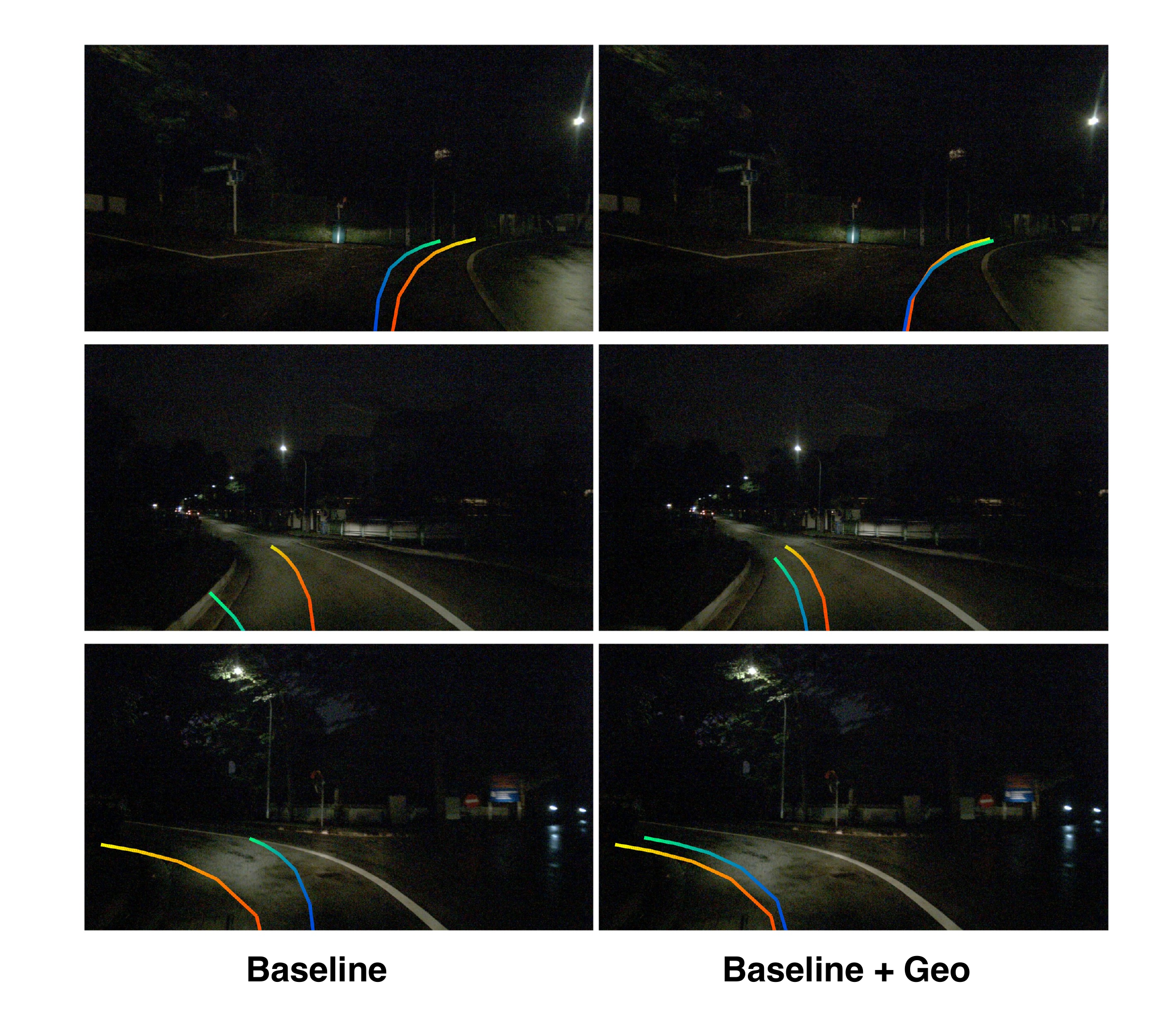}
    \caption{
        \textbf{Planning:} The figure compares the planning performance between the baseline(VAD) and our Spatial Augmented method in night scenes. 
        While the baseline struggles driving in low light, our method acts as a robust guide, generating safer and more consistent trajectories. 
    }
    \label{fig:planning}
\end{figure}

\begin{figure}[]
    \centering
    \includegraphics[width=\linewidth]{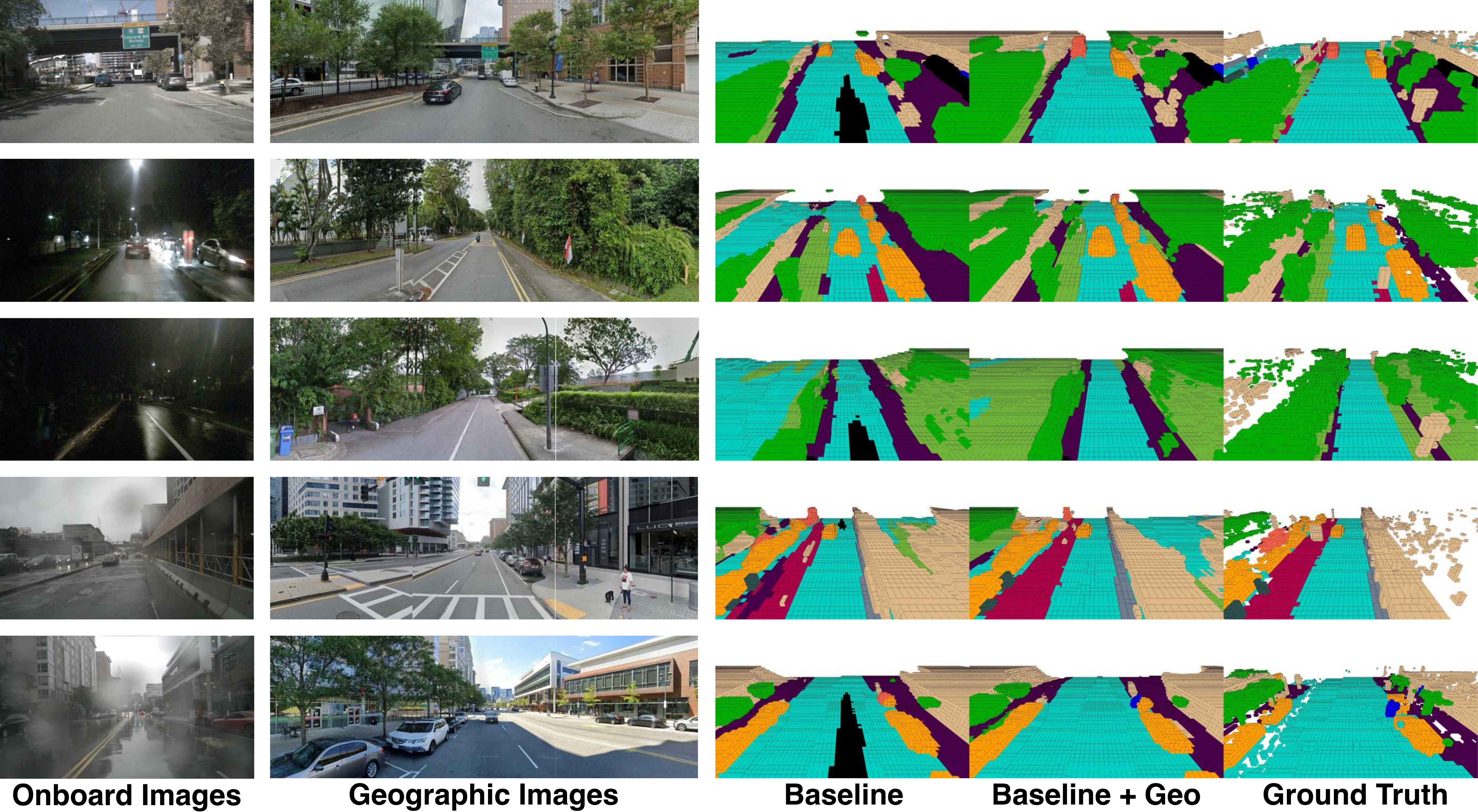}
    \caption{
        \textbf{Occupancy Prediction.} Our method, benefiting from the spatial priors provided by geographic images, produces more complete and coherent occupancy estimations, particularly under challenging conditions such as adverse weather, motion blur, and occlusions.
    }
    \label{fig:supp-occ-case}
\end{figure}

\begin{figure}[]
    \centering
    \includegraphics[width=\linewidth]{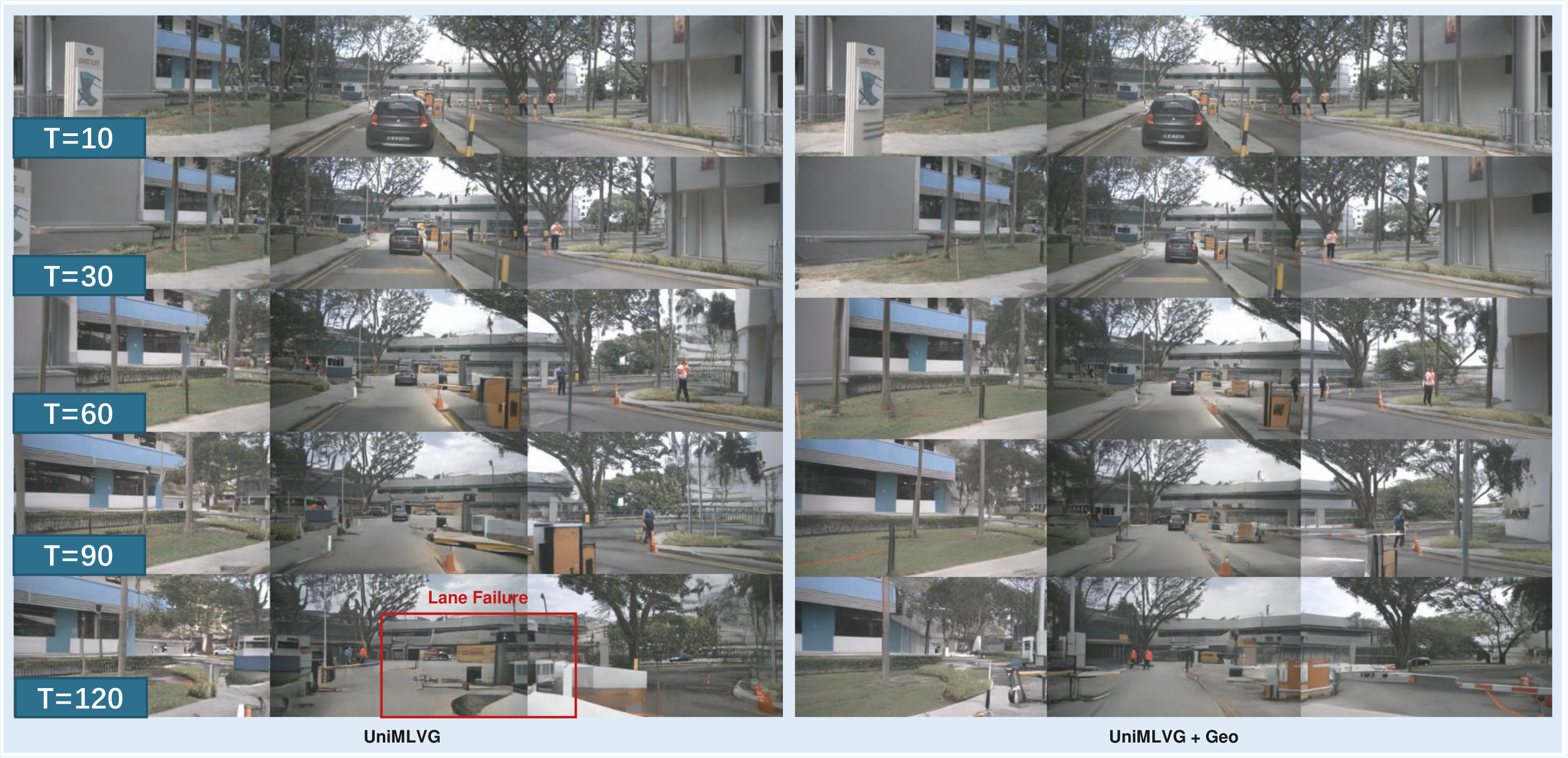}
    \caption{
        \textbf{Generative World Model.} In the presence of geographic images, the autoregressive generation becomes more stable.
    }
    \label{fig:gen_case2}
\end{figure}

We provide additional qualitative examples to demonstrate the effectiveness of our approach in challenging scenarios. The online mapping results are shown in Fig.~\ref{fig:map_case1} with Fig.~\ref{fig:map_case2} ,the occupancy prediction results are shown in Fig.~\ref{fig:supp-occ-case}, the planning results are shown in Fig.~\ref{fig:planning} and the generation results in Fig.~\ref{fig:gen_case2}.

\clearpage

{
    \small
    \bibliographystyle{ieeenat_fullname}
    \bibliography{supp}
}


\makeatletter\let\@\cvpr@origat\makeatother

\makeatletter
\let\maketitle\cvpr@origmaketitle
\let\bibliography\cvpr@origbibliography
\let\bibliographystyle\cvpr@origbibliographystyle
\let\tableofcontents\cvpr@origtoc
\let\section\cvpr@origsection
\let\subsection\cvpr@origsubsection
\let\subsubsection\cvpr@origsubsubsection
\makeatother

\clearpage
{\small
\bibliographystyle{ieeenat_fullname}
\bibliography{supp}
}
\end{document}